\documentclass[journal]{IEEEtran}
\usepackage{amsmath,amsfonts,amssymb}
\usepackage[linesnumbered,ruled]{algorithm2e}
\usepackage{amsmath}
\usepackage{array}
\usepackage[caption=false,font=normalsize,labelfont=sf,textfont=sf]{subfig}
\usepackage{textcomp}
\usepackage{stfloats}
\usepackage{url}
\usepackage{verbatim}
\usepackage{graphicx}
\usepackage{bm}
\usepackage{cite}
\usepackage{multirow}
\usepackage{color}
\usepackage{mathtools}
\usepackage{amsthm}

\DeclareMathOperator*{\argmax}{argmax}

\usepackage[table]{xcolor}
\newcommand{\best}[1]{\cellcolor{gray!18}\textbf{#1}}

\begin{document}

\title{In-Context Guidance: Learning Inter-Task Synergies via Numerical Foundational Models for Few-Shot Multitask Optimization}

% \author{Anonymous Authors}

\author{Tingyang Wei,
        Haofeng Wu,\IEEEmembership{~Member,~IEEE},
        Jiao Liu, \IEEEmembership{~Member,~IEEE},
        Zhao Wei,
        Puay Siew Tan,
        and~Yew-Soon Ong,\IEEEmembership{~Fellow,~IEEE}% <-this % stops a space
        \thanks{T. Wei, H. Wu and J. Liu are with the College of Computing and Data Science, Nanyang Technological University, Singapore (e-mail: TINGYANG001@e.ntu.edu.sg, haofeng.wu@ntu.edu.sg, jiao.liu@ntu.edu.sg)}% <-this % stops a space
        \thanks{Z. Wei is with the Institute of Advanced Intelligence and Computing, Agency for Science, Technology and Research, Singapore (e-mail: weiz@a-star.edu.sg)}
        
        \thanks{P. S. Tan is with the Singapore Institute of Manufacturing
        Technology (SIMTech), Agency for Science, Technology and Research, Singapore (e-mail: pstan@simtech.a-star.edu.sg)}% <-this % stops a space
        \thanks{Y.-S. Ong is with the College of Computing and Data Science, Nanyang Technological University, Singapore, and also with the Centre for Frontier AI Research, A*STAR, and the Institute of Advanced Intelligence and Computing, Singapore (e-mail: asysong@ntu.edu.sg)}
}
\markboth{Journal of \LaTeX\ Class Files,~Vol.~14, No.~8, August~2021}%
{Shell \MakeLowercase{\textit{et al.}}: A Sample Article Using IEEEtran.cls for IEEE Journals}

\maketitle
\begin{abstract}
Multi-task optimization (MTO) addresses a set of optimization tasks simultaneously, often suffering from inaccurate inter-task relationship estimation under limited evaluation budgets, leading to negative transfer. 
This paper introduces In-Context Guidance Multitask Optimization (ICG-MTO), a novel framework that leverages numerical foundational models to improve inter-task coupling estimation in few-shot scenarios. 
Unlike conventional methods that rely solely on scarce observed data, ICG-MTO employs a frozen foundational model to infer auxiliary guidance through in-context learning. 
The framework operates through three stages: constructing an algorithm-specific in-context query from evaluated solutions, using the foundational model to infer a guidance signal characterizing predictive relationships among tasks, and translating this signal into algorithm-specific guidance for maximum-a-posteriori coupling estimation.
This approach provides regularization during the early, data-scarce stages of optimization and gradually relinquishes control as task-specific observations accumulate.  
We instantiate the framework in multitask Bayesian optimization as ICG-MTBO, using directional fitness-class queries to guide inter-task coupling estimation, and further instantiate it in MFEA-II using decision-space-overlap queries to guide random mating probability estimation.
Experiments across synthetic benchmarks and a real-world robot arm control problem, together with evaluations under different acquisition functions and evolutionary multitasking, demonstrate the effectiveness and generality of ICG-MTO for few-shot multitask optimization.
\end{abstract}


\section{Introduction}

\IEEEPARstart{M}{ultitask} optimization aims to solve multiple optimization problems simultaneously by exploiting inter-task synergy~\cite{wei, PMTO}. 
Given a predefined set of optimization tasks, the key challenge is not merely whether knowledge should be transferred~\cite{mfea, mfea2}, but how strongly the search processes across tasks should be coupled so that beneficial transfer can be promoted while negative transfer is mitigated under limited evaluation budgets.
For instance, the pioneering multifactorial evolutionary algorithm~(MFEA)~\cite{mfea} employs a random mating probability, $rmp$, to control the intensity of inter-task transfer. 
The follow-up work, MFEA-II~\cite{mfea2}, introduces an adaptive mixture-model based estimator to control this transfer according to the evolving task populations. 
More generally, multitask optimizers contain an explicit or implicit \emph{inter-task coupling mechanism} that determines how observations~\cite{tuituji, selection-framework, adaptive, curbing, mfea2} obtained from one task influence the search on another.

Estimation of such inter-task coupling is particularly difficult under few-shot settings. Let $\mathcal{D}_k$ denote the evaluated solutions collected from task $\mathcal{T}_k$, and let $\boldsymbol{\vartheta}$ denote a set of generic inter-task coupling parameters. A conventional data-driven estimator can be written as
\begin{equation}
\widehat{\boldsymbol{\vartheta}}
=
\argmax_{\boldsymbol{\vartheta}}
\mathcal{L}
\left(
\boldsymbol{\vartheta};
\mathcal{D}_1,\ldots,\mathcal{D}_K
\right),
\label{eq:ch3-canonical-coupling-estimation}
\end{equation}
where $\mathcal{L}$ denotes an estimator-specific objective, such as a marginal likelihood~\cite{MTBO}, a mixture likelihood~\cite{mfea2}, or a fitness-based measure of transfer utility~\cite{gmm3}. During the early stage of optimization, when knowledge transfer is potentially most valuable, the limited datasets $\{\mathcal{D}_k\}_{k=1}^{K}$ may provide insufficient evidence for reliably estimating $\boldsymbol{\vartheta}$, leading to inaccurate inter-task coupling and potentially negative transfer~\cite{parting, mfea2}. This challenge arises across different multitask optimizers: multitask Bayesian optimization estimates the inter-task coupling matrix of a multitask Gaussian process from limited observations~\cite{MTBO}, while evolutionary multitasking estimates random mating probabilities~\cite{mfea, mfea2} or other transfer-related parameters~\cite{tuituji} from small evolving populations.

This motivates the use of an auxiliary source of information to guide inter-task coupling estimation during the data-scarce stage of optimization. Numerical foundational models provide one such source: pretrained on broad distributions of synthetic numerical or tabular datasets, they can perform in-context prediction on small context datasets through a forward pass, without task-specific fine-tuning or additional function evaluations~\cite{tabpfn,tfn_01,tfn_02}. Building on this capability, we introduce \emph{In-Context Guidance Multitask Optimization}~(ICG-MTO), which uses a frozen numerical foundational model to infer auxiliary guidance for inter-task coupling estimation. Rather than directly using its predictions as the coupling estimate, ICG-MTO incorporates the inferred guidance as a prior over the algorithm-specific coupling parameters:
\begin{equation}
\widehat{\boldsymbol{\vartheta}}_{\mathrm{MAP}}
=
\argmax_{\boldsymbol{\vartheta}}
\left[
\mathcal{L}
\left(
\boldsymbol{\vartheta};
\mathcal{D}_1,\ldots,\mathcal{D}_K
\right)
+
\log
p
\left(
\boldsymbol{\vartheta}
\mid
\boldsymbol{\vartheta}^*
\right)
\right],
\label{eq:ch3-map-general}
\end{equation}
where $\boldsymbol{\vartheta}^*$ denotes the algorithm-specific reference coupling parameters derived from the inferred in-context guidance.

The ICG-MTO framework follows three stages.
First, an \emph{in-context query} is constructed by reorganizing the evaluated solutions into a supervised prediction problem aligned with the inter-task coupling mechanism of the base multitask optimizer.
Second, the frozen numerical foundational model performs in-context prediction on the constructed queries and produces an in-context guidance signal characterizing the predictive relationships among tasks.
Third, this signal is translated into algorithm-specific guidance and incorporated into the inter-task coupling estimator through a maximum-a-posteriori formulation, as shown in \eqref{eq:ch3-map-general}.
The corresponding guidance weight is largest when evaluated solutions are scarce and is gradually annealed as task-specific observations accumulate.
The framework therefore supports the inter-task coupling estimator during its least reliable stage while allowing the base multitask optimizer to progressively fall back to its conventional data-driven behavior.
We instantiate this framework in two representative multitask optimizers: ICG-MTBO, which uses directional fitness-class queries to guide the inter-task coupling estimation of a multitask Bayesian optimizer, and ICG-MFEA-II, which uses decision-space-overlap queries to guide the estimation of the random mating probability matrix in MFEA-II~\cite{mfea2}.

The main contributions of this paper are summarized as follows:
\begin{itemize}
    \item An \emph{In-Context Guidance Multitask Optimization} framework is proposed to improve inter-task coupling estimation under limited evaluation budgets. It uses a frozen numerical foundational model as an auxiliary source of guidance without replacing the underlying multitask optimizer.

    \item An in-context query construction strategy is developed for multitask Bayesian optimization. It transforms evaluated solutions into directional fitness-class prediction queries, where task-specific inter-task guidance can be inferred.

    \item A guidance-informed maximum-a-posteriori estimator is developed for the multitask Gaussian process. The inferred in-context guidance is incorporated as an annealed regularization term over the inter-task coupling matrix, whose influence gradually diminishes as optimization observations accumulate.

    \item The framework is evaluated with two different acquisition functions and is further instantiated in MFEA-II through a \emph{decision-space-overlap query}, demonstrating that the proposed in-context guidance principle is not restricted to a particular acquisition strategy or multitask optimization paradigm.
\end{itemize}

\section{Preliminaries}
\label{sec:ch3-preliminaries}

\subsection{Multitask Bayesian Optimization}
\label{subsec:ch3-mtbo}

Multitask Bayesian optimization~\cite{MTBO} extends conventional Bayesian optimization by jointly optimizing multiple related black-box optimization tasks.
Rather than constructing an independent surrogate model for each task, multitask Bayesian optimization employs a multitask Gaussian process~(MTGP)~\cite{MTGP} to model the objective functions jointly over the decision space and the task space.
By explicitly parameterizing inter-task coupling, the MTGP allows observations collected from one task to improve the predictive model of another related task, thereby facilitating knowledge transfer under limited evaluation budgets.

The MTGP instantiated in this paper follows the intrinsic coregionalization model~(ICM)~\cite{MTBO}, which assumes a separable covariance structure:
\begin{equation}
    \kappa_{\mathrm{mt}}
    \left(
        (\mathbf{x},k),
        (\mathbf{x}',k')
    \right)
    =
    \kappa_{\mathbf{x}}
    \left(
        \mathbf{x},\mathbf{x}'
    \right)
    \cdot
    \mathbf{B}[k,k'],
    \label{eq:ch3-separable-kernel}
\end{equation}
where $\kappa_{\mathbf{x}}(\mathbf{x},\mathbf{x}')$ measures the similarity between two solutions in the decision space, while $\mathbf{B}\in\mathbb{S}_{+}^{K}$ denotes the positive semidefinite inter-task coupling matrix for the $K$ optimization tasks.
Under this formulation, the covariance between two observations is jointly determined by their similarity in the decision space and the coupling between their corresponding tasks.
The matrix $\mathbf{B}$ therefore governs how information is shared across tasks and directly affects the intensity and quality of knowledge transfer within the MTGP.

In this paper, the kernel function in the decision space is defined using an automatic relevance determination~(ARD) radial basis function~(RBF) kernel~\cite{williams2006gaussian}:
\begin{equation}
    \kappa_{\mathbf{x}}
    \left(
        \mathbf{x},\mathbf{x}'
    \right)
    =
    \exp
    \left(
        -\frac{1}{2}
        \sum_{i=1}^{V}
        \frac{
            (x_i-x'_i)^2
        }{
            \ell_i^2
        }
    \right),
    \label{eq:ch3-ard-rbf}
\end{equation}
where $\ell_i$ denotes the length-scale parameter associated with the $i$-th decision variable.
The ARD formulation assigns a distinct length scale to each dimension, allowing the surrogate model to characterize the relative importance of individual decision variables~\cite{williams2006gaussian}.

Under the \texttt{Positive Index Kernel} in Gpytorch~\cite{Gpytorch}, the inter-task coupling matrix is parameterized as
\begin{equation}
    \mathbf{B}
    =
    \mathbf{L}\mathbf{L}^{\top}
    +
    \operatorname{diag}(\mathbf{v}),
    \label{eq:ch3-positive-index-kernel}
\end{equation}
where $\mathbf{L}$ is a non-negative lower-triangular matrix and $\mathbf{v}$ is a non-negative vector.
This parameterization ensures that $\mathbf{B}$ is positive semidefinite and therefore defines a valid task covariance matrix.
Because non-negativity is imposed on both $\mathbf{L}$ and $\mathbf{v}$, the resulting task covariances are restricted to be non-negative.

Let $\boldsymbol{\vartheta}$ collect the inter-task coupling parameters, the decision-space kernel parameters, and the observation-noise parameters.
Given the observations $\mathbf{y}$, the corresponding inputs $\mathbf{X}$, and the covariance matrix $\mathbf{K}_{\boldsymbol{\vartheta}}$, conventional MTGP training estimates $\boldsymbol{\vartheta}$ through maximum likelihood estimation:
\begin{align}
    \widehat{\boldsymbol{\vartheta}}_{\mathrm{MLE}}
    &=
    \argmax_{\boldsymbol{\vartheta}}
    \log p
    \left(
        \mathbf{y}
        \mid
        \mathbf{X},
        \boldsymbol{\vartheta}
    \right)
    \nonumber\\
    &=
    \argmax_{\boldsymbol{\vartheta}}
    \left[
        -\frac{1}{2}
        \mathbf{y}^{\top}
        \mathbf{K}_{\boldsymbol{\vartheta}}^{-1}
        \mathbf{y}
        -
        \frac{1}{2}
        \log
        \left|
            \mathbf{K}_{\boldsymbol{\vartheta}}
        \right|
        -
        \frac{n}{2}
        \log(2\pi)
    \right].
    \label{eq:ch3-mtgp-mle}
\end{align}

Maximum likelihood estimation relies solely on the collected observations to infer the task coupling matrix and the remaining kernel hyperparameters~\cite{MTBO}.
Its quality therefore depends on whether the available data provide sufficient evidence.
Under few-shot settings, the marginal likelihood may be weakly informative such that multiple inter-task coupling structures explain the limited observations similarly well~\cite{ill_posed}.
As a result, maximum likelihood estimators with limited samples can be susceptible to instability or overfitting, while GP kernel hyperparameter estimation can become ill-posed under particular model and observation conditions~\cite{bishop2006pattern,proof_mle,penalized_mle,ill_posed}.

A Bayesian estimation perspective provides a natural means of regularizing this estimation process.
Instead of estimating the parameters solely from the marginal likelihood, maximum a posteriori~(MAP) estimation combines the evidence provided by the observations with additional prior information~\cite{bishop2006pattern,bayesian_mle}:
\begin{equation}
    \widehat{\boldsymbol{\vartheta}}_{\mathrm{MAP}}
    =
    \argmax_{\boldsymbol{\vartheta}}
    \left[
        \log p
        \left(
            \mathbf{y}
            \mid
            \mathbf{X},
            \boldsymbol{\vartheta}
        \right)
        +
        \log p
        \left(
            \boldsymbol{\vartheta}
        \right)
    \right].
    \label{eq:ch3-mtgp-map}
\end{equation}
The prior term regularizes the estimator by favoring plausible parameter configurations when the likelihood provides insufficient evidence.
Such regularization can improve parameter estimation under limited observations by discouraging configurations that are weakly supported by the collected data~\cite{bishop2006pattern,proof_mle,penalized_mle}.

Since negative transfer in the MTGP is governed primarily by the inter-task coupling matrix, the central question under few-shot settings is how to construct an informative guidance term for estimating $\mathbf{B}$ in \eqref{eq:ch3-separable-kernel}.
A generic ad hoc prior may not adequately represent the heterogeneous inter-task relationships arising across different optimization problems.
This motivates the construction of an auxiliary, data-dependent guidance signal that complements the limited likelihood information and regularizes the estimation of the inter-task coupling matrix.
In this paper, such a guidance signal is inferred through in-context learning using a frozen numerical foundational model. 
It is subsequently incorporated into the MTGP through a guidance-informed MAP estimator.

\subsection{Numerical Foundational Models and In-Context Learning}
\label{subsec:ch3-pfn}

Recent advances in foundational models have demonstrated that large-scale pretraining can produce general-purpose learners that adapt to new tasks through in-context learning~\cite{bommasani2021opportunities}.
While this paradigm was initially established in language and vision using large text and image corpora, an emerging direction considers \emph{numerical foundational models}, whose pretraining data consist of large collections of synthetic numerical datasets~\cite{tabpfn}.
These models acquire inductive biases over numerical relationships, functional structures, and predictive procedures that can be transferred to previously unseen tabular learning problems~\cite{tabpfn}.

A popular numerical foundational model is Prior-Data Fitted Network~(PFN)~\cite{bayesian_transformers}, which reformulates Bayesian posterior prediction as a supervised learning problem over datasets.
Given a prior distribution over supervised learning tasks, PFNs are pretrained on a large collection of synthetic datasets sampled from that prior and learn to approximate Bayesian inference.
Formally, given a context dataset
\begin{equation}
\mathcal{C}
=
\left\{
(\mathbf{x}^{(i)},y^{(i)})
\right\}_{i=1}^{N},
\label{eqn:context}
\end{equation}
a PFN approximates the posterior predictive distribution
\begin{equation}
    \operatorname{PFN}
    \left(
        y
        \mid
        \mathbf{x},
        \mathcal{C}
    \right)
    \approx
    p
    \left(
        y
        \mid
        \mathbf{x},
        \mathcal{C}
    \right)
    =
    \int
    p
    \left(
        y
        \mid
        \mathbf{x},
        f
    \right)
    p
    \left(
        f
        \mid
        \mathcal{C}
    \right)
    \mathrm{d}f,
    \label{eq:ch3-pfn-predictive}
\end{equation}
where $f$ denotes the latent predictive function.
Unlike conventional Bayesian inference, which performs posterior computation separately for every new dataset, the transformer amortizes this inference during pretraining and enables posterior prediction through a forward pass~\cite{tabpfn,bayesian_transformers,tfn_01,tfn_02}.

TabPFN~\cite{tabpfn} is a practical instantiation of PFNs designed for tabular learning under small-data cases.
This pretrained model can function as a general-purpose predictor that is competitive across diverse tabular datasets without task-specific fine-tuning~\cite{tabpfn}.
The adaptation behavior of TabPFN can be attributed to \emph{in-context learning}~(ICL)~\cite{ICL}.
Instead of updating model parameters, ICL conditions the frozen transformer on a context dataset and performs prediction directly from the provided examples.
Let $\mathcal{C}$ in \eqref{eqn:context} denote the context set, and let $\mathbf{x}_q$ denote a query sample.
The prediction from ICL can be obtained as
\begin{equation}
    \widehat{y}_q
    =
    \mathcal{F}
    \left(
        \mathbf{x}_q,
        \mathcal{C}
    \right),
    \label{eq:ch3-icl-prediction}
\end{equation}
where $\mathcal{F}$ denotes the pretrained transformer with fixed parameters.
The model therefore adapts to a new prediction query entirely by conditioning on the context examples rather than through gradient-based fine-tuning.

In this paper, the numerical foundational model is not used to approximate the objective functions directly or replace the multitask Gaussian process. Instead, it serves as an auxiliary guidance module for improving inter-task coupling estimation from limited observations. To align its prediction task with the inter-task coupling mechanism of the base multitask optimizer, the evaluated solutions are reorganized into algorithm-specific \emph{in-context queries}. The frozen numerical foundational model then performs in-context prediction on these queries to infer guidance signals characterizing inter-task relationships. These signals are subsequently translated into algorithm-specific reference coupling parameters and incorporated into the coupling estimator through a guidance-informed prior.

\begin{figure*}[!t]
    \centering
    \includegraphics[width=0.90\textwidth]{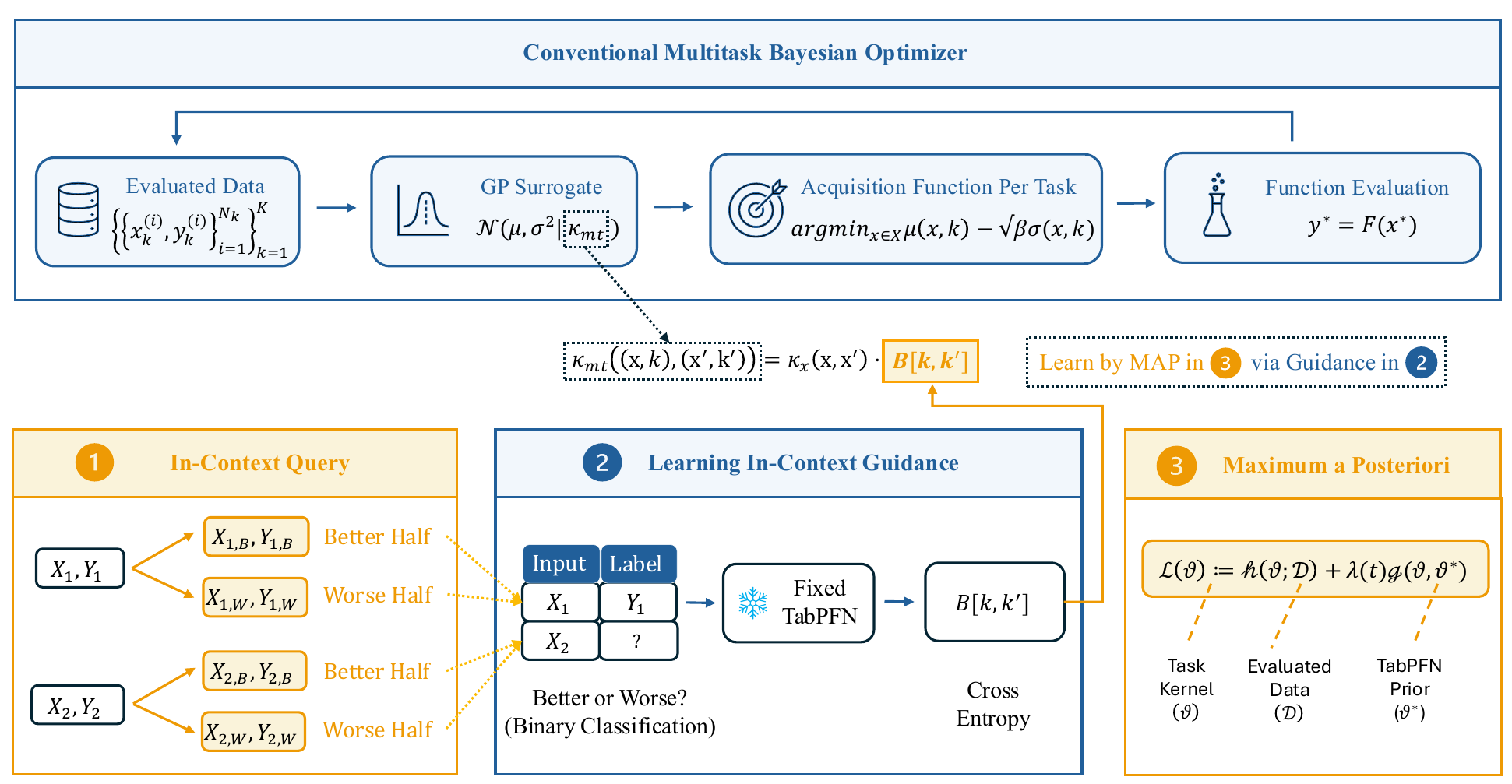}
    \caption{Overall workflow of the proposed In-Context Guidance Multitask Bayesian Optimization (ICG-MTBO) framework. The upper pipeline corresponds to a conventional multitask Bayesian optimizer, where evaluated solutions are used to construct an MTGP surrogate for acquisition optimization and subsequent function evaluation. The lower pipeline augments inter-task coupling estimation through three sequential stages. (1) \emph{In-context query} reorganizes the evaluated solutions into algorithm-specific prediction queries by partitioning the observations of each task into representative fitness classes (the better and worse halves in ICG-MTBO). (2) \emph{In-context guidance learning} employs a frozen numerical foundational model (TabPFN) to perform cross-task prediction on the constructed queries, producing guidance signals that characterize predictive relationships among tasks. (3) \emph{Maximum-a-posteriori} estimation translates the inferred guidance into reference coupling parameters and incorporates them as a prior for estimating the MTGP task coupling matrix. The resulting task kernel is used by the MTGP surrogate, while the remaining components of the Bayesian optimization framework remain unchanged.}
    \label{fig:ch3-overview}
\end{figure*}

\section{In-Context Guidance Multitask Bayesian Optimization}
\label{sec: ch3-bayesian}

\subsection{Overview}
\label{subsec:ch3-overview}

The proposed In-Context Guidance Multitask Bayesian Optimization (ICG-MTBO) augments conventional multitask Bayesian optimization with a numerical foundational model that provides auxiliary guidance for inter-task coupling estimation under limited evaluation budgets. Unlike approaches that replace the surrogate model~\cite{meta_llm} or modify the acquisition strategy~\cite{llm_acqf}, ICG-MTBO preserves the conventional multitask Bayesian optimization pipeline and intervenes only in the estimation of the inter-task coupling matrix of the multitask Gaussian process (MTGP). The MTGP surrogate, uncertainty quantification, acquisition optimization, and objective evaluation therefore remain unchanged.

Figure~\ref{fig:ch3-overview} illustrates the overall workflow. Evaluated solutions are used both by the conventional MTGP estimator and by the proposed guidance mechanism, which consists of three stages:
\begin{itemize}
    \item First, \emph{in-context query} reorganizes the evaluated solutions into algorithm-specific prediction queries aligned with the inter-task coupling mechanism of MTBO.
    \item Second, \emph{in-context guidance learning} utilizes a frozen numerical foundational model to perform cross-task prediction on the constructed queries and infer guidance signals characterizing predictive relationships among tasks. 
    \item Finally, \emph{MAP estimation} translates the inferred guidance into reference coupling parameters and incorporates them as a prior for estimating the MTGP inter-task coupling matrix. The resulting task kernel is subsequently used by the conventional MTGP surrogate for acquisition optimization and function evaluation.
\end{itemize}

ICG-MTBO therefore maintains a separation between \emph{optimization} and \emph{guidance}: the multitask Bayesian optimizer remains responsible for objective modeling, uncertainty quantification, and candidate selection, while the numerical foundational model provides auxiliary information for inter-task coupling estimation. This modularity also enables the same ICG principle to be instantiated with other multitask optimizers, as demonstrated later with MFEA-II~\cite{mfea2}.

\subsection{In-Context Query Construction}
\label{subsec:ch3-curriculum}

To align the prediction task of the numerical foundational model with inter-task coupling estimation, the evaluated solutions are reorganized into directional fitness-class prediction queries. Consider task $\mathcal{T}_k$ with $N_k$ evaluated solutions,
\begin{equation}
    \mathcal{D}_{k}
    =
    \left\{
        \left(
            \mathbf{x}_{k}^{(i)},
            y_{k}^{(i)}
        \right)
    \right\}_{i=1}^{N_k},
    \label{eq:ch3-task-dataset}
\end{equation}
where $\mathbf{x}_{k}^{(i)}\in\mathcal{X}$ and $y_{k}^{(i)}$ denote the evaluated solution and its objective value, respectively. Since objective values may have different scales and distributions across tasks, each observation is first assigned a task-wise rank,
\begin{equation}
    r_{k}^{(i)}
    =
    \operatorname{rank}
    \left(
        y_{k}^{(i)};
        \{y_{k}^{(j)}\}_{j=1}^{N_k}
    \right),
    \label{eq:ch3-fitness-rank}
\end{equation}
and subsequently mapped into $C$ ordinal fitness classes:
\begin{equation}
    \ell_{k}^{(i)}
    =
    \operatorname{clip}
    \left(
        \left\lfloor
            \frac{r_{k}^{(i)}C}{N_k}
        \right\rfloor,
        0,
        C-1
    \right).
    \label{eq:ch3-fitness-class}
\end{equation}
In this paper, $C=2$ is adopted, partitioning the observations of each task into the \emph{better} and \emph{worse} halves according to their objective values. The resulting classified dataset is
\begin{equation}
    \widetilde{\mathcal{D}}_{k}
    =
    \left\{
        \left(
            \mathbf{x}_{k}^{(i)},
            \ell_{k}^{(i)}
        \right)
    \right\}_{i=1}^{N_k}.
    \label{eq:ch3-classified-dataset}
\end{equation}

For each ordered task pair $(k_1,k_2)$, an in-context query is then constructed as
\begin{equation}
    \mathcal{Q}_{k_1\rightarrow k_2}
    =
    \left(
        \widetilde{\mathcal{D}}_{k_1},
        \{
            \mathbf{x}_{k_2}^{(i)}
        \}_{i=1}^{N_{k_2}}
    \right),
    \label{eq:ch3-personalized-query}
\end{equation}
where the classified observations from $\mathcal{T}_{k_1}$ provide the in-context examples and the evaluated solutions from $\mathcal{T}_{k_2}$ serve as query inputs. The query therefore examines whether the fitness structure observed on $\mathcal{T}_{k_1}$ can predict the better and worse regions of $\mathcal{T}_{k_2}$. Since predictive utility can be asymmetric across tasks~\cite{dra,mto-dra}, $\mathcal{Q}_{k_1\rightarrow k_2}$ and $\mathcal{Q}_{k_2\rightarrow k_1}$ are treated as distinct directional queries. These queries are subsequently passed to the frozen numerical foundational model to infer inter-task guidance.

\subsection{Learning In-Context Guidance}
\label{subsec:ch3-icl-guidance}

Given the directional in-context queries constructed in Section~\ref{subsec:ch3-curriculum}, the frozen TabPFN is used to infer predictive relationships among tasks without task-specific fine-tuning or parameter updates. For each ordered task pair $(k_1,k_2)$, the classified observations of $\mathcal{T}_{k_1}$ provide the in-context examples, while the evaluated solutions of $\mathcal{T}_{k_2}$ serve as prediction inputs. For each target solution $\mathbf{x}_{k_2}^{(i)}$, TabPFN produces a categorical predictive distribution:
\begin{equation}
    \mathbf{p}_{k_1\rightarrow k_2}^{(i)}
    =
    \mathcal{F}
    \left(
        \ell
        \mid
        \mathbf{x}_{k_2}^{(i)};
        \widetilde{\mathcal{D}}_{k_1}
    \right),
    i=1,\ldots,N_{k_2},
    \label{eq:ch3-tabpfn-prediction}
\end{equation}
where
\begin{equation}
    \mathbf{p}_{k_1\rightarrow k_2}^{(i)}
    \in
    [0,1]^C,
    \sum_{c=0}^{C-1}
    p_{k_1\rightarrow k_2,c}^{(i)}
    =
    1.
\end{equation}
The predictive utility from source task $\mathcal{T}_{k_1}$ to target task $\mathcal{T}_{k_2}$ is quantified by the mean cross-entropy with respect to the true fitness-class labels:
\begin{equation}
    \operatorname{CE}[k_1,k_2]
    =
    -
    \frac{1}{N_{k_2}}
    \sum_{i=1}^{N_{k_2}}
    \log
    \mathcal{F}
    \left(
        \ell_{k_2}^{(i)}
        \mid
        \mathbf{x}_{k_2}^{(i)};
        \widetilde{\mathcal{D}}_{k_1}
    \right).
    \label{eq:ch3-directed-ce}
\end{equation}
A smaller $\operatorname{CE}[k_1,k_2]$ indicates that the fitness structure observed on $\mathcal{T}_{k_1}$ is more informative for distinguishing the better and worse regions of $\mathcal{T}_{k_2}$. Since the source and target tasks play different roles in the prediction, this utility is directional.

To obtain a bounded guidance signal, the cross-entropy is calibrated against the random-classification baseline $\log C$:
\begin{equation}
    \Delta[k_1,k_2]
    =
    \operatorname{clip}
    \left(
        \log C
        -
        \operatorname{CE}[k_1,k_2],
        0,
        \log C
    \right).
    \label{eq:ch3-ce-improvement}
\end{equation}
The directed guidance is then defined as
\begin{equation}
    S[k_1,k_2]
    =
    \left(
        \frac{
            \Delta[k_1,k_2]
        }{
            \log C
        }
    \right)^{1/\tau},
    S[k,k]=1,
    \label{eq:ch3-similarity-calibration}
\end{equation}
where $\tau>0$ controls the calibration sharpness and is set to $\tau=1$ in the experiments. Thus, $S[k_1,k_2]\in[0,1]$, with predictions no better than random mapped to zero and perfect cross-task classification mapped to one. Applying this procedure to all ordered task pairs yields
\begin{equation}
    \mathbf{S}
    =
    \left[
        S[k_1,k_2]
    \right]_{k_1,k_2=1}^{K}
    \in
    [0,1]^{K\times K}.
    \label{eq:ch3-directed-similarity-matrix}
\end{equation}
Here, $S[k_1,k_2]$ measures the predictive utility of $\mathcal{T}_{k_1}$ for $\mathcal{T}_{k_2}$, and $\mathbf{S}$ is therefore not necessarily symmetric.

To translate this directional guidance into a form suitable for MTGP coupling estimation, a target-specific symmetric correlation guidance matrix
\begin{equation}
    \mathbf{R}_k
    \in
    \mathbb{R}^{K\times K}
\end{equation}
is constructed for each target task $\mathcal{T}_k$. Its correlations involving the target task retain the corresponding directed guidance:
\begin{equation}
    R_k[j,k]
    =
    R_k[k,j]
    =
    S[j,k],
    \qquad
    j\neq k.
    \label{eq:ch3-target-correlation}
\end{equation}
For two non-target tasks, both transfer directions are combined using their geometric mean:
\begin{equation}
    R_k[j,q]
    =
    R_k[q,j]
    =
    \sqrt{
        S[j,q]S[q,j]
    },
    \qquad
    j,q\neq k.
    \label{eq:ch3-source-correlation}
\end{equation}
The diagonal entries are fixed to one:
\begin{equation}
    R_k[j,j]
    =
    1,
    \qquad
    j=1,\ldots,K.
    \label{eq:ch3-correlation-diagonal}
\end{equation}
The complete construction is therefore
\begin{equation}
    R_k[j,q]
    =
    \begin{cases}
        S[j,k],
        & q=k,\; j\neq k,
        \\[1mm]
        S[q,k],
        & j=k,\; q\neq k,
        \\[1mm]
        \sqrt{S[j,q]S[q,j]},
        & j,q\neq k,\; j\neq q,
        \\[1mm]
        1,
        & j=q.
    \end{cases}
    \label{eq:ch3-directed-guidance-matrix}
\end{equation}
Since $\mathbf{R}_k$ is not necessarily positive semidefinite, it is projected onto the positive semidefinite cone before being used with the MTGP:
\begin{equation}
    \mathbf{R}_k
    \leftarrow
    \operatorname{make\_psd}
    \left(
        \mathbf{R}_k;
        \varepsilon
    \right),
    \qquad
    \varepsilon=10^{-4},
    \label{eq:ch3-make-psd}
\end{equation}
where eigenvalues smaller than $\varepsilon$ are clipped before reconstructing the matrix~\cite{Gpytorch}. The resulting $\mathbf{R}_k$ preserves target-specific directional guidance while satisfying the requirements of a valid task kernel. Importantly, $\mathbf{R}_k$ is not used directly as the MTGP coupling matrix; it provides reference guidance for estimating the inter-task coupling matrix in \eqref{eq:ch3-positive-index-kernel}.

\subsection{Guidance Adaptation through MAP Estimation}
\label{subsec:ch3-map}

The target-specific correlation guidance matrix $\mathbf{R}_k$ obtained in Section~\ref{subsec:ch3-icl-guidance} is not directly used as the final task coupling matrix of the MTGP. Instead, it defines a reference coupling structure
\begin{equation}
    \mathbf{B}_{\mathrm{guidance},k}
    =
    \mathbf{R}_k.
    \label{eq:ch3-guidance-coupling}
\end{equation}
For each target task $\mathcal{T}_k$, a separate MTGP is fitted using observations across all tasks, and the guidance is incorporated through a prior over its inter-task coupling matrix. Specifically, we assume
\begin{equation}
    \operatorname{vec}(\mathbf{B})
    \sim
    \mathcal{N}
    \left(
        \operatorname{vec}
        \left(
            \mathbf{B}_{\mathrm{guidance},k}
        \right),
        \tau_{B}^{2}\mathbf{I}
    \right),
    \label{eq:ch3-b-prior}
\end{equation}
where $\tau_{B}^{2}$ represents the uncertainty associated with the inferred guidance. The corresponding negative log-prior, up to a constant independent of $\mathbf{B}$, is
\begin{equation}
    -\log p(\mathbf{B})
    =
    \frac{1}{2\tau_{B}^{2}}
    \left\|
        \mathbf{B}
        -
        \mathbf{B}_{\mathrm{guidance},k}
    \right\|^{2}
    + \mathrm{const}.
    \label{eq:ch3-negative-log-prior}
\end{equation}

Let $\boldsymbol{\vartheta}$ collect the coupling parameters, decision-space kernel parameters, and observation-noise parameters of the MTGP. Combining the conventional marginal likelihood with the guidance-informed prior gives the MAP objective
\begin{equation}
    \mathcal{J}_k
    \left(
        \boldsymbol{\vartheta}
    \right)
    =
    -
    \log
    p
    \left(
        \mathbf{y}
        \mid
        \mathbf{X},
        \boldsymbol{\vartheta}
    \right)
    +
    \lambda(t)
    \left\|
        \mathbf{B}
        \left(
            \boldsymbol{\vartheta}
        \right)
        -
        \mathbf{B}_{\mathrm{guidance},k}
    \right\|^{2},
    \label{eq:ch3-map-objective}
\end{equation}
where $\mathbf{B}(\boldsymbol{\vartheta})$ denotes the coupling matrix induced by the current MTGP parameters and
\begin{equation}
    \lambda
    =
    \frac{1}{2\tau_{B}^{2}}.
    \label{eq:ch3-lambda-prior-variance}
\end{equation}
Thus, the guidance complements rather than replaces conventional marginal-likelihood estimation. A larger $\lambda$ places greater emphasis on the reference coupling structure, whereas a smaller $\lambda$ allows the observed optimization data to dominate the estimation~\cite{bishop2006pattern}. The remaining MTGP parameters continue to be estimated conventionally from the observations.

\subsubsection{Theoretical Inspiration}
The strength of the guidance can be interpreted heuristically through a pseudo-observation analogy. Let $\sigma^{2}$ denote an effective observation variance associated with the coupling estimator~\cite{neuenschwander2020predictively}. The relative prior precision then corresponds to approximately
\begin{equation}
    n_{0}
    =
    \frac{\sigma^{2}}{\tau_{B}^{2}}
    =
    2\sigma^{2}\lambda
    \label{eq:ch3-pseudo-observations}
\end{equation}
virtual observations~\cite{neuenschwander2020predictively}. This quantity characterizes the relative information contributed by the guidance rather than an exact effective sample size of the complete MTGP. Under few-shot observations, such supplementary information can stabilize coupling estimation; as observations accumulate, the marginal likelihood provides increasingly informative statistical evidence~\cite{proof_mle,bayesian_mle,penalized_mle}.

\subsubsection{Annealed Guidance Weight}
Accordingly, the guidance weight is annealed throughout optimization:
\begin{equation}
    \lambda(t)
    =
    \lambda_{0}
    \exp
    \left(
        -\delta t
    \right),
    \qquad
    \lambda_{0}=1.0,
    \qquad
    \delta=0.05,
    \label{eq:ch3-lambda-schedule}
\end{equation}
where $t$ denotes the optimization iteration and $\delta$ controls the decay rate. Under the pseudo-observation interpretation, the corresponding guidance strength becomes
\begin{equation}
    n_{0}(t)
    =
    2\sigma^{2}\lambda(t)
    =
    2\sigma^{2}\lambda_{0}
    \exp
    \left(
        -\delta t
    \right)
    \longrightarrow
    0
    \quad
    \text{as}
    \quad
    t\longrightarrow\infty.
    \label{eq:ch3-pseudo-observation-decay}
\end{equation}
Hence, the guidance has its strongest influence during the early data-scarce stage and progressively diminishes as observations accumulate. In the limit, $\lambda(t)\rightarrow 0$ and $n_0(t)\rightarrow 0$, such that the estimation of the inter-task coupling matrix falls back to the conventional marginal-likelihood-based MTGP estimator. This annealing mechanism allows the in-context guidance to regularize coupling estimation when observations are scarce without permanently constraining the task relationships learned from the optimization data.

\subsection{Guided Multitask Bayesian Optimization}
\label{subsec:ch3-guided-bo}

Once the task coupling matrix has been estimated through the MAP formulation, the resultant MTGP is applied in the conventional Bayesian optimization pipeline. 
Let $\mu_k(\mathbf{x})$ and $\sigma_k(\mathbf{x})$ denote the posterior mean and posterior standard deviation of the MTGP for task $\mathcal{T}_k$ at solution $\mathbf{x}$. 
These posterior quantities are then passed to a standard acquisition function to determine the next candidate solution.

\subsubsection{Lower Confidence Bound}

The main experiments employ a lower confidence bound~(LCB) acquisition function~\cite{andrea}. 
On the normalized objective scale used by the implementation, the acquisition value for task $\mathcal{T}_m$ is defined as
\begin{equation}
    \alpha_{\mathrm{LCB}}
    \left(
        \mathbf{x};
        \mathcal{T}_k
    \right)
    =
    \mu_k
    \left(
        \mathbf{x}
    \right)
    -
    \sqrt{\beta}
    \,
    \sigma_k
    \left(
        \mathbf{x}
    \right)
    \label{eq:ch3-lcb}
\end{equation}
where $\beta$ controls the trade-off between exploitation and exploration.
The acquisition function therefore favors solutions with either a promising posterior mean or a high predictive uncertainty.

\subsubsection{Overall ICG-MTBO procedure}

Algorithm~\ref{alg:ch3-icg-mtbo} summarizes the complete ICG-MTBO procedure. 
At each optimization iteration, the collected observations are first transformed into the in-context queries. 
The frozen TabPFN then performs cross-task in-context learning to generate in-context guidance signals. 
These values are translated into target-specific reference coupling matrices, which define guidance-informed priors for the annealed MAP estimation of the MTGP inter-task coupling matrices.
Finally, a conventional acquisition function selects the next candidate solution for each active target task.

\begin{algorithm}[t]
\caption{In-Context Guidance Multitask Bayesian Optimization}
\label{alg:ch3-icg-mtbo}

\KwIn{
Optimization tasks $\{\mathcal{T}_k\}_{k=1}^{K}$; evaluation budget $N$ per task; number of fitness classes $C$; initial guidance strength $\lambda_{0}$; decay rate $\delta$; calibration parameter $\tau$; trade-off parameter $\beta$.
}
\KwOut{
Best solution for each optimization task.
}

Initialize each dataset $\mathcal{D}_k$ using evaluated Latin-hypercube samples\;
Set optimization iteration $t\leftarrow0$\;

\While{$\sum_{k=1}^{K}N_k<KN$}{
    Identify the active task set
    $\mathcal{A}\leftarrow
    \{
        \mathcal{T}_k:
        N_k<N
    \}$\;

    Compute the annealed guidance weight
    $\lambda(t)\leftarrow
    \lambda_{0}\exp(-\delta t)$\;

    Normalize the collected objective values\;

    \tcp{Step 1: In-Context Query Construction}
    \ForEach{task $\mathcal{T}_k$}{
        Construct fitness-class labels using~\eqref{eq:ch3-fitness-class}\;
        Form the classified dataset $\widetilde{\mathcal{D}}_k$\;
    }

    \tcp{Step 2: Learning In-Context Guidance}
    \ForEach{ordered task pair $(k_1,k_2)$ with $k_1\neq k_2$}{
        Present $\widetilde{\mathcal{D}}_{k_1}$ as context to the frozen TabPFN\;
        Predict the fitness classes of task $\mathcal{T}_{k_2}$\;
        Compute $\operatorname{CE}[k_1,k_2]$ using~\eqref{eq:ch3-directed-ce}\;
        Calibrate $S[k_1,k_2]$ using~\eqref{eq:ch3-similarity-calibration}\;
    }

    \ForEach{active target task $\mathcal{T}_k\in\mathcal{A}$}{
        Construct the target-specific correlation guidance matrix
        $\mathbf{R}_k$ using~\eqref{eq:ch3-directed-guidance-matrix}\;

        Project $\mathbf{R}_k$ onto the positive-semidefinite cone\;

        \tcp{Step 3: Guidance adaptation through MAP Estimation}
        Initialize a fresh MTGP using the observations shared across all tasks\;

        Initialize the task kernel from $\mathbf{R}_k$ and obtain
        $\mathbf{B}_{\mathrm{guidance},k}$\;

        Fit the MTGP by minimizing~\eqref{eq:ch3-map-objective}\;

        \tcp{Step 4: Guided Multitask Bayesian optimization}
        Optimize
        $\alpha_{\mathrm{LCB}}$ in~\eqref{eq:ch3-lcb}
        for task $\mathcal{T}_k$\;

        Evaluate the selected solution and update $\mathcal{D}_k$\;
    }

    Update $t\leftarrow t+1$\;
}

\Return{The best solution from each $\mathcal{D}_k$}\;
\end{algorithm}

%========================================================
% Convergence trends on 30-dimensional synthetic problems
%========================================================

\subsubsection{Computational Complexity}
The dominant computational cost of each optimization iteration stems from fitting a single MTGP for each active target task. 
The in-context guidance module additionally requires $K(K-1)$ TabPFN forward queries to evaluate all ordered task pairs. 
These computations do not consume the evaluation budget and are assumed to be substantially less costly than evaluating the underlying black-box objective functions in the few-shot optimization settings.

\section{Results}
To evaluate the effectiveness of the proposed algorithm, we compare it with representative baseline methods on both synthetic benchmark problems and a real-world application. The baselines include the single-task Bayesian optimizer~(BO)~\cite{aq-total, Human-Loop}, the multitask Bayesian optimizer~(MTBO)~\cite{MTBO}, and several state-of-the-art few-shot multitask optimizers, namely BO-LCB-CKT~\cite{xue_ckt}, BO-LCB-BCKT~\cite{xue_bckt}, and the evolutionary multitask Bayesian optimizer SELF~\cite{SELF}. By default, the LCB acquisition function is adopted for all methods to ensure consistency with the publicly available implementations of BO-LCB-CKT, BO-LCB-BCKT, and SELF~\cite{xue_ckt, xue_bckt, SELF}.

To further assess the generality of the conclusions, we also compare the proposed method with STBO and MTBO using the acquisition function, LogEI~\cite{LogEI}, a numerically stable reformulation of Expected Improvement~\cite{expected_improvement}. 
It is worth noting that the proposed framework is not restricted to a specific multitask optimizer. 
In the methodology, we instantiate ICG-MTBO by introducing in-context guidance into the estimation of the MTBO task coupling matrix. 
To further demonstrate the generality of the framework, we additionally develop a variant based on the well-established evolutionary multitask optimizer MFEA-II~\cite{mfea2}, in which in-context guidance is applied to estimate the random mating probability~(rmp) matrix and thereby improve its multitask search performance. 
Finally, the stability of the proposed method is examined through a sensitivity analysis.

\subsection{Benchmark Definitions}
\subsubsection{Synthetic Benchmarks}
We first conduct comparative studies on the single-objective multitask benchmark suite~\cite{single-benchmark}. 
It comprises nine multitask optimization problems, each containing two tasks whose relationship is characterized along two features, that is, the intersection of their global optima and the similarity of their search spaces. 
We evaluate two different versions of synthetic problems.
We evaluate the 50-dimensional problems as originally defined, and a 30-dimensional counterpart in
which each task is restricted to the first $30$ dimensional variables of the full $50$ dimensional landscape that passes through the global optimum.
Detailed settings can be found in Section S-I in supplementary materials.
\subsubsection{Real Problems}
\begin{figure}[!t]
    \centering    \includegraphics[width=0.27\textwidth]{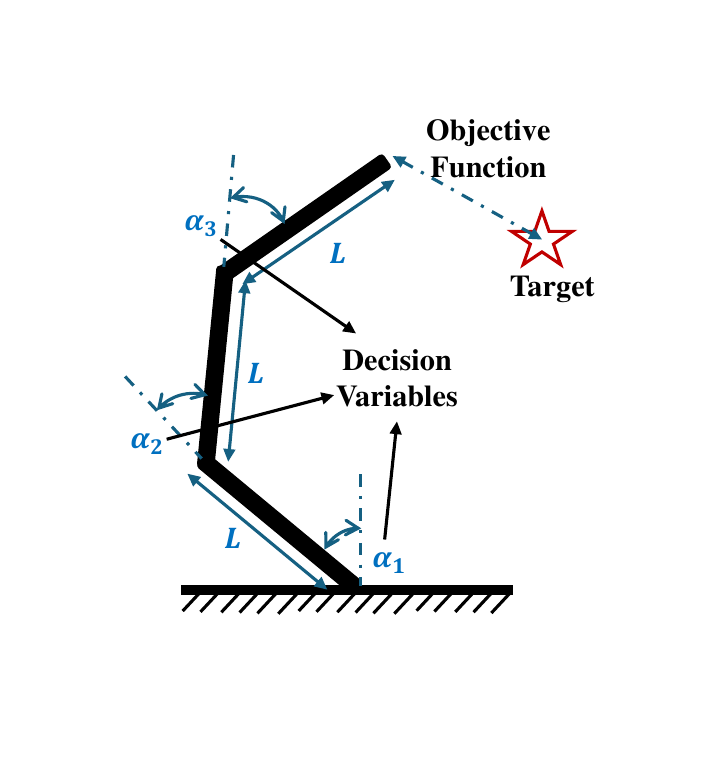}
    \caption{Illustration of the separable robot-arm reaching problem (SepArm). The decision variables determine the normalized angular commands of the joints, which are mapped to the corresponding physical joint angles. The objective is to minimize the distance between the end-effector position and a fixed target. Across the multitask problem bundle, the tasks share the same target and total arm length but differ in the allowable angular range of the joints, resulting in different levels of inter-task similarity.}
    \label{fig: ch3_separm}
\end{figure}

\begin{table*}[t]
\centering
\caption{Statistical summary of the comparative results on the 30-dimensional synthetic benchmark problems. The $+/-/\approx$ entries report the numbers of tasks on which each method performs significantly better than, significantly worse than, or statistically similarly to ICG-MTBO, respectively, according to the Wilcoxon test at a significance level of $0.05$. The best average rank at each evaluation budget is highlighted in bold with a shaded background.}
\label{tab:ch3-synthetic-30d-summary}

\resizebox{0.70\textwidth}{!}{%
\begin{tabular}{c|c|c|c|c|c|c|c}
\hline
Metric
& Evaluations
& BO-LCB
& MTBO-LCB
& BO-LCB-CKT
& BO-LCB-BCKT
& SELF
& ICG-MTBO
\\
\hline\hline

\multirow{4}{*}{$+/-/\approx$}
& 40
& 0/16/2
& 0/14/4
& 4/12/2
& 0/17/1
& 4/6/8
& -
\\
\cline{2-8}

& 60
& 0/15/3
& 1/13/4
& 3/13/2
& 0/17/1
& 1/7/10
& -
\\
\cline{2-8}

& 80
& 0/16/2
& 2/9/7
& 1/15/2
& 0/18/0
& 0/14/4
& -
\\
\cline{2-8}

& 100
& 0/15/3
& 2/8/8
& 1/15/2
& 0/18/0
& 0/14/4
& -
\\
\hline\hline

\multirow{4}{*}{Average rank}
& 40
& 4.94
& 3.11
& 3.44
& 5.11
& 2.50
& \best{1.89}
\\
\cline{2-8}

& 60
& 5.06
& 2.78
& 3.94
& 5.00
& 2.61
& \best{1.61}
\\
\cline{2-8}

& 80
& 4.83
& 2.39
& 4.28
& 5.06
& 2.89
& \best{1.56}
\\
\cline{2-8}

& 100
& 4.78
& 2.17
& 4.17
& 5.06
& 3.28
& \best{1.56}
\\
\hline\hline

\end{tabular}%
}
\end{table*}

\begin{table*}[t]
\centering
\caption{Statistical summary of the comparative results on the 50-dimensional synthetic benchmark problems. The $+/-/\approx$ entries report the numbers of tasks on which each method performs significantly better than, significantly worse than, or statistically similarly to ICG-MTBO, respectively, according to the Wilcoxon test at a significance level of $0.05$. The best average rank at each evaluation budget is highlighted in bold with a shaded background.}
\label{tab:ch3-synthetic-50d-summary}

\resizebox{0.70\textwidth}{!}{%
\begin{tabular}{c|c|c|c|c|c|c|c}
\hline
Metric
& Evaluations
& BO-LCB
& MTBO-LCB
& BO-LCB-CKT
& BO-LCB-BCKT
& SELF
& ICG-MTBO
\\
\hline\hline

\multirow{4}{*}{$+/-/\approx$}
& 40
& 0/14/4
& 0/13/5
& 0/15/3
& 0/15/3
& 0/7/11
& -
\\
\cline{2-8}

& 60
& 0/15/3
& 1/15/2
& 0/15/3
& 0/15/3
& 3/6/9
& -
\\
\cline{2-8}

& 80
& 0/15/3
& 0/14/4
& 0/15/3
& 0/17/1
& 1/10/7
& -
\\
\cline{2-8}

& 100
& 0/17/1
& 0/15/3
& 0/15/3
& 0/17/1
& 1/14/3
& -
\\
\hline\hline

\multirow{4}{*}{Average rank}
& 40
& 4.22
& 3.00
& 4.56
& 4.83
& 2.72
& \best{1.67}
\\
\cline{2-8}

& 60
& 4.17
& 3.11
& 4.78
& 5.17
& 1.94
& \best{1.83}
\\
\cline{2-8}

& 80
& 4.44
& 3.00
& 4.61
& 5.28
& 2.17
& \best{1.50}
\\
\cline{2-8}

& 100
& 4.39
& 2.56
& 4.78
& 5.28
& 2.61
& \best{1.39}
\\
\hline\hline

\end{tabular}%
}
\end{table*}

For the real-world study, we adopt the separable robot-arm reaching problem (SepArm)~\cite{qd-multitask}, in which the angular positions of the joints are adjusted so that the end effector reaches a fixed target as closely as possible\footnote{One can refer to \cite{qd-multitask} for the exact forward-kinematics formulation.}. 
As shown in Fig.~\ref{fig: ch3_separm}, the solution is the vector of normalized joint commands $\mathbf{x}\in[0,1]^{V}$, each entry mapped to a physical angle in $[-\pi a_{\max}, \pi a_{\max}]$, and the objective quantifies the distance between the end effector position and the target.
We construct three sets of problems where each set represents high similarity~(HS), medium similarity~(MS), and low similarity~(LS) multitask problems across 5-D, 10-D, and 15-D search spaces, yielding nine problems P1-P9: 5-D HS/MS/LS (P1-P3), 10-D HS/MS/LS (P4-P6), and 15-D HS/MS/LS (P7-P9).
More details can be found in Section S-I in our supplementary materials.

\subsection{Experimental Setup}
All experiments follow the few-shot multitask settings, where each task is generally allocated a total budget of $100$ function evaluations. 
For the synthetic benchmarks, every optimizer is initialized with $20$ samples per task generated by Latin hypercube sampling~(LHS)~\cite{lhs}, leaving $80$ iterative evaluations for the optimization phase.
However, for the real SepArm problems, a smaller initial design of $5$ samples per task is adopted for the lower-dimensional problems, and the total budget for each task is set to $60$ function evaluations. 
All methods are assigned an identical evaluation budget, and each experiment is independently repeated $20$ times.
Performance is reported as the best objective value found so far, and the convergence curves show the mean across all runs, with a shaded band representing a standard deviation.

For the compared GP-based methods (BO, MTBO, BO-LCB-CKT, BO-LCB-BCKT, and SELF), the
surrogate is a Gaussian process with an RBF ARD kernel, the inputs
are normalized to $[0,1]^{V}$, the objectives are min-max normalized, and the hyperparameters are fitted by maximizing the exact marginal likelihood.
The acquisition function is optimized by multi-start gradient ascent with $5$ restarts and $200$ steps per restart. 
The LCB exploration coefficient is
set to $\beta=2.5$ for the GP baselines.
The remaining hyperparameters of the baselines follow their original publications.
For the proposed ICG-MTBO, the in-context guidance is produced by TabPFN~v2.5~\cite{grinsztajn2025tabpfn}, a tabular foundation model pretrained on synthetic datasets, accessed through the
official implementation (\texttt{tabpfn} package, version 7.0.1)
No fine-tuning is performed, and a single forward pass is utilized per estimate.
The MAP estimate of the task coupling matrix is obtained with an initial regularization weight
$\lambda_{0}=1$ that decays exponentially at rate $0.05$ per iteration, so that the in-context guidance dominates in the data-scarce early stage and gradually falls back to the likelihood as observations accumulate.

One can refer to Section S-I in supplementary materials for the full details of generality study on acquisition function and MFEA-II.

\subsection{Discussions}
\subsubsection{Competitive Optimization Performance}
Tables~S-I--S-IV show that ICG-MTBO achieves consistently competitive performance across both the 30- and 50-dimensional benchmark suites.
Summarized results can be found in Table~I--II.
In particular, it obtains the best average rank at every evaluation budget, with ranks improving from 1.89 to 1.56 in the 30-dimensional case and from 1.67 to 1.39 in the 50-dimensional case, showcasing that in higher-dimensional cases where the evaluation budgets are more limited, ours can achieve a larger performance margin. 
Although several competing methods outperform ICG-MTBO on a small number of individual tasks, the proposed method is significantly better than or statistically comparable to them in the majority of comparisons, demonstrating stable performance across different problem dimensions and stages of the optimization process.

% Required packages:
% \usepackage{rotating}
% \usepackage{multirow}
% \usepackage{graphicx}

% Required packages:
% \usepackage{rotating}
% \usepackage{multirow}
% \usepackage{graphicx}

\subsubsection{The Effectiveness of In-Context Guidance}
A direct comparison between ICG-MTBO and its unguided counterpart, MTBO-LCB, provides an ablation study of the proposed in-context guidance mechanism. 
Across the 30-dimensional benchmarks, ICG-MTBO performs better than, statistically similarly to, and worse than MTBO-LCB in 44, 23, and 5 out of 72 comparisons, respectively. 
The advantage becomes more pronounced in the 50-dimensional setting, where the corresponding counts are 57, 14, and 1. 
Overall, across both dimensions and all evaluation budgets, ICG-MTBO achieves better, comparable, and worse performance in 101, 37, and 6 out of 144 comparisons, respectively. 
Notably, at the earliest budget of 40 evaluations, ICG-MTBO is better in 27 comparisons and comparable in the remaining 9, without being significantly worse in any case. 
This confirms that the ICG component is particularly effective during the early few-shot stage, when the conventional likelihood-based estimate of inter-task coupling remains unreliable. As the optimization proceeds, especially in the lower-dimensional setting, the performance gap gradually narrows, and the proposed method increasingly becomes statistically comparable to MTBO-LCB. 
This trend is consistent with the annealing design, where the influence of the in-context guidance is progressively reduced, and the optimizer gradually falls back toward the behavior of standard MTBO-LCB once sufficient observations have been accumulated.

Nevertheless, the occasional inferior results also reveal the limitation of the guidance mechanism. 
The fitness-class query in ICG-MTBO provides only a coarse representation of the task landscapes, and the resultant in-context guidance may not accurately characterize complex or weakly transferable inter-task relationships. 
Consequently, inaccurate in-context guidance can temporarily bias the estimated task coupling matrix.
The overall results therefore suggest that the main benefit of ICG lies in stabilizing and improving inter-task relationship estimation during the early few-shot stage, while its influence appropriately diminishes as the data-driven estimator of MTBO-LCB becomes more reliable.

\begin{table}[t]
\centering
\caption{Statistical summary of the comparative results using the LogEI acquisition function on the 50-dimensional synthetic benchmark problems. The $+/-/\approx$ entries report the numbers of tasks on which each method performs significantly better than, significantly worse than, or statistically similarly to ICG-MTBO, respectively, according to the Wilcoxon test at a significance level of $0.05$. The best average rank at each evaluation budget is highlighted in bold with a shaded background.}
\label{tab:ch3-synthetic-50d-logei-summary}

\resizebox{0.47\textwidth}{!}{%
\begin{tabular}{c|c|c|c|c}
\hline
Metric
& Evaluations
& BO-LogEI
& MTBO-LogEI
& ICG-MTBO-LogEI
\\
\hline\hline

\multirow{4}{*}{$+/-/\approx$}
& 40
& 2/13/3
& 1/11/6
& -
\\
\cline{2-5}

& 60
& 1/14/3
& 1/13/4
& -
\\
\cline{2-5}

& 80
& 1/14/3
& 1/14/3
& -
\\
\cline{2-5}

& 100
& 0/14/4
& 2/12/4
& -
\\
\hline\hline

\multirow{4}{*}{Average rank}
& 40
& 2.50
& 2.11
& \best{1.39}
\\
\cline{2-5}

& 60
& 2.61
& 2.11
& \best{1.28}
\\
\cline{2-5}

& 80
& 2.56
& 2.17
& \best{1.28}
\\
\cline{2-5}

& 100
& 2.67
& 1.94
& \best{1.39}
\\
\hline\hline

\end{tabular}%
}
\end{table}

\begin{table}[t]
\centering
\caption{Statistical summary of the comparative results using the LogEI acquisition function on the 30-dimensional synthetic benchmark problems. The $+/-/\approx$ entries report the numbers of tasks on which each method performs significantly better than, significantly worse than, or statistically similarly to ICG-MTBO, respectively, according to the Wilcoxon test at a significance level of $0.05$. The best average rank at each evaluation budget is highlighted in bold with a shaded background.}
\label{tab:ch3-synthetic-30d-logei-summary}

\resizebox{0.47\textwidth}{!}{%
\begin{tabular}{c|c|c|c|c}
\hline
Metric
& Evaluations
& BO-LogEI
& MTBO-LogEI
& ICG-MTBO-LogEI
\\
\hline\hline

\multirow{4}{*}{$+/-/\approx$}
& 40
& 0/15/3
& 1/10/7
& -
\\
\cline{2-5}

& 60
& 1/14/3
& 2/11/5
& -
\\
\cline{2-5}

& 80
& 1/14/3
& 3/9/6
& -
\\
\cline{2-5}

& 100
& 2/10/6
& 3/7/8
& -
\\
\hline\hline

\multirow{4}{*}{Average rank}
& 40
& 2.89
& 1.94
& \best{1.17}
\\
\cline{2-5}

& 60
& 2.78
& 2.00
& \best{1.22}
\\
\cline{2-5}

& 80
& 2.72
& 1.83
& \best{1.44}
\\
\cline{2-5}

& 100
& 2.56
& 1.78
& \best{1.67}
\\
\hline\hline

\end{tabular}%
}
\end{table}

\begin{figure*}[h]
    \centering

    % Row 1: P2 and P4
    \subfloat[P2 (CIMS)]{
        \includegraphics[width=0.48\textwidth]{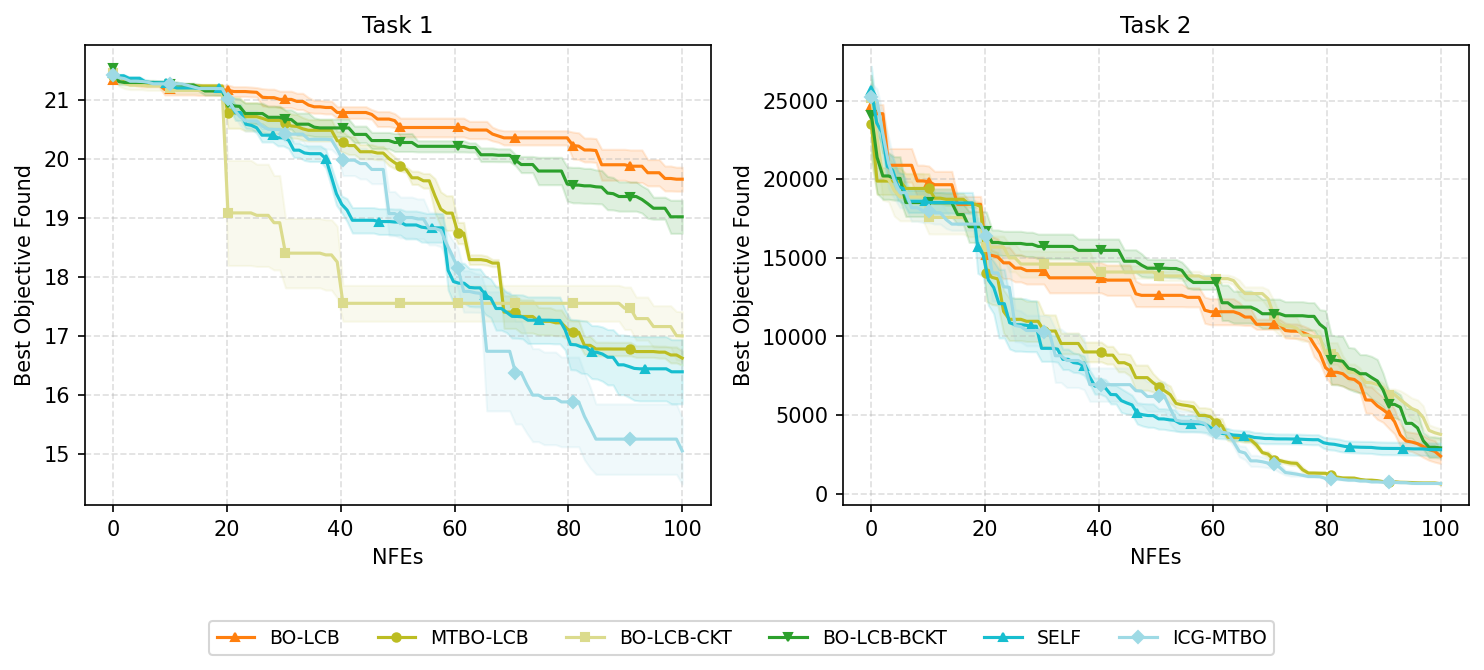}
        \label{fig:ch3-convergence-30d-p1}
    }
    \hfill
    \subfloat[P4 (PIHS)]{
        \includegraphics[width=0.48\textwidth]{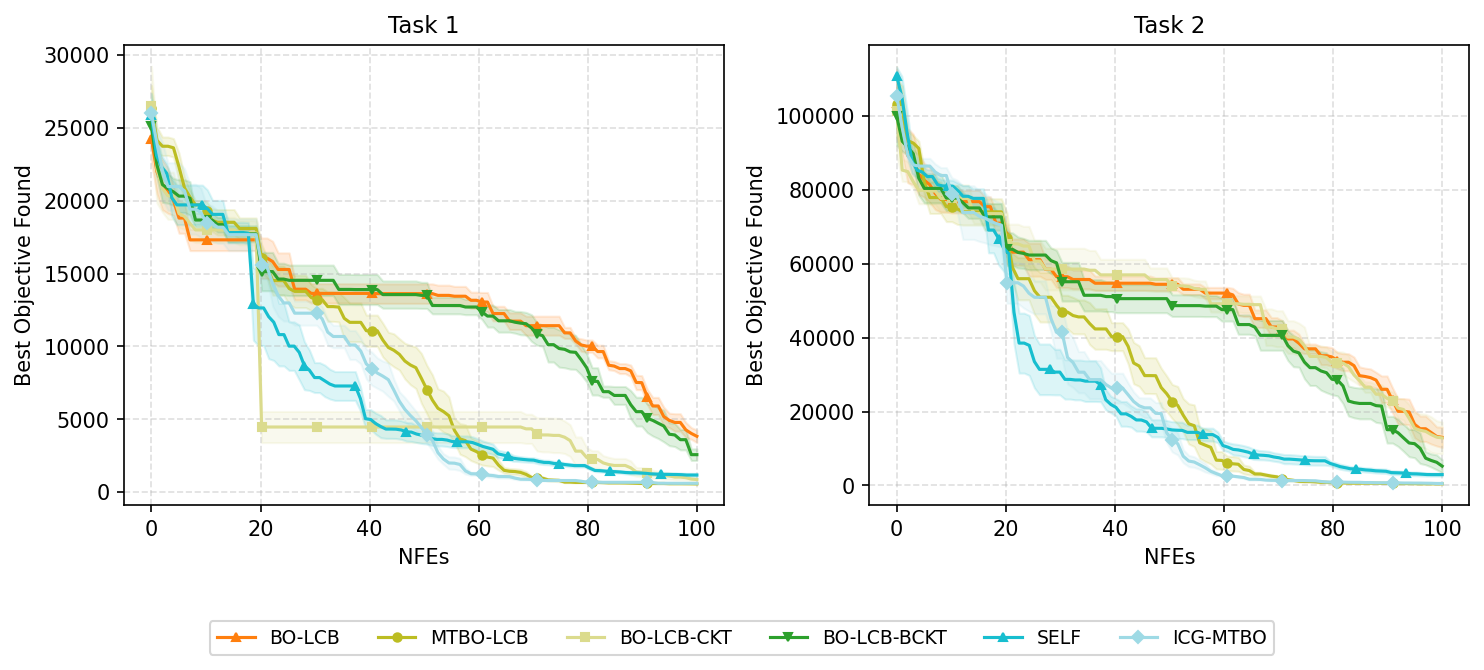}
        \label{fig:ch3-convergence-30d-p2}
    }

    \vspace{0.5mm}

    % Row 2: P6 and P8
    \subfloat[P6 (PILS)]{
        \includegraphics[width=0.48\textwidth]{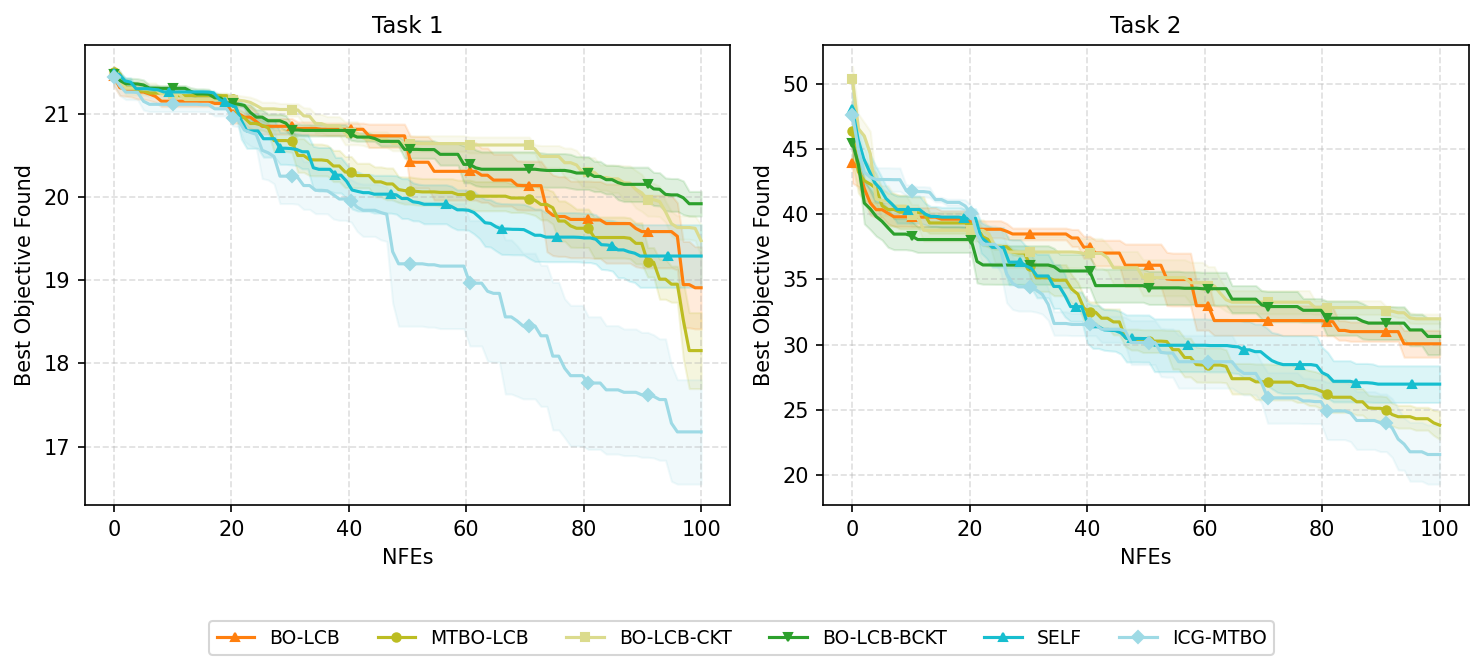}
        \label{fig:ch3-convergence-30d-p3}
    }
    \hfill
    \subfloat[P8 (NIMS)]{
        \includegraphics[width=0.48\textwidth]{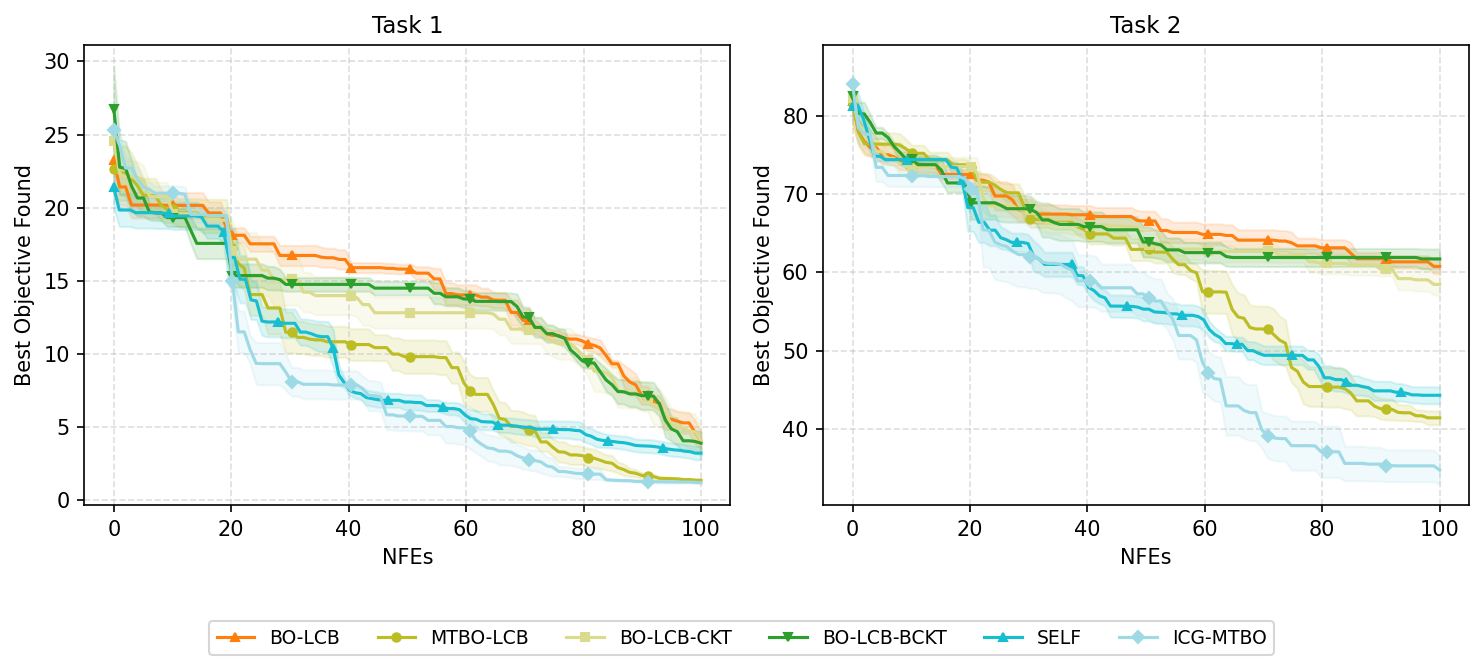}
        \label{fig:ch3-convergence-30d-p4}
    }

    \caption{Convergence trends of BO-LCB, MTBO-LCB, BO-LCB-CKT, BO-LCB-BCKT, SELF, and ICG-MTBO on the 30-dimensional synthetic benchmark problems.
    (a) P2 (CIMS).
    (b) P4 (PIHS).
    (c) P6 (PILS).
    (d) P8 (NIMS).}
    \label{fig:ch3-convergence-30d}
\end{figure*}

\begin{figure*}[h]
    \centering

    % Row 1: P1 and P6
    \subfloat[P1 (CIHS)]{
        \includegraphics[width=0.48\textwidth]{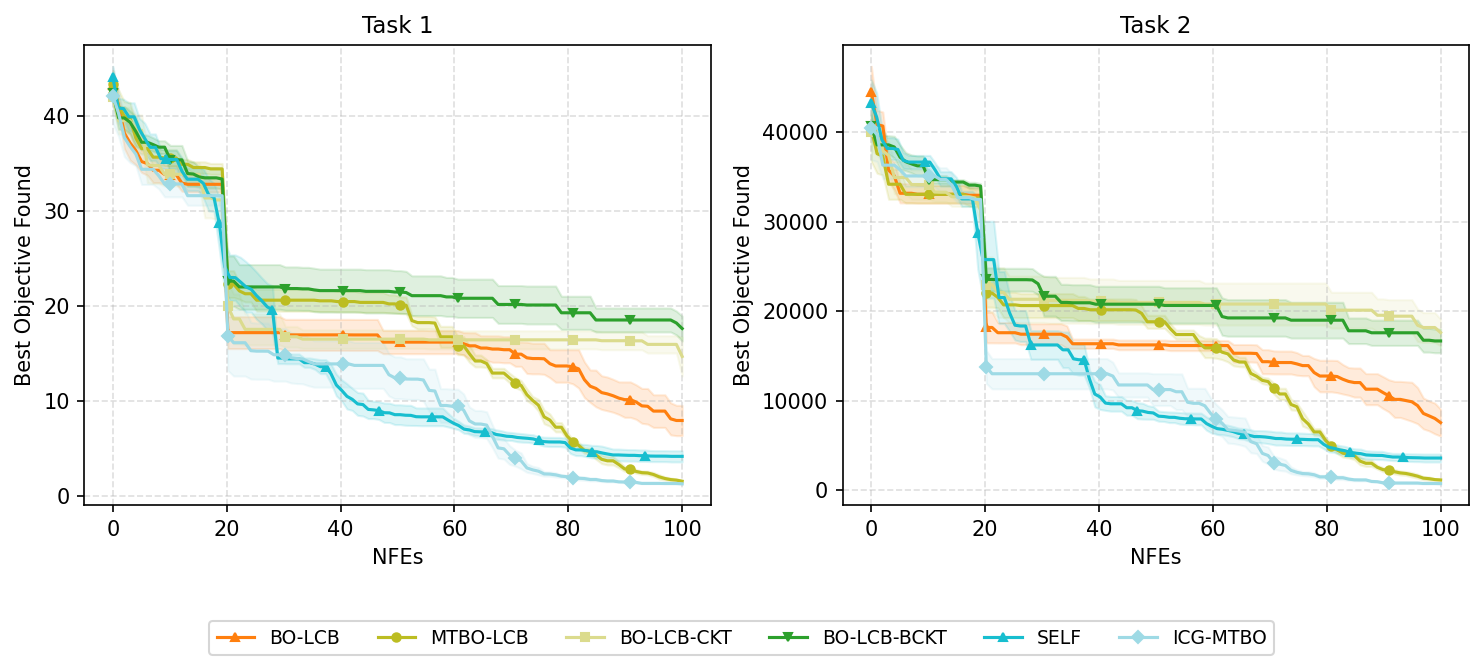}
        \label{fig:ch3-convergence-50d-p1}
    }
    \hfill
    \subfloat[P6 (PILS)]{
        \includegraphics[width=0.48\textwidth]{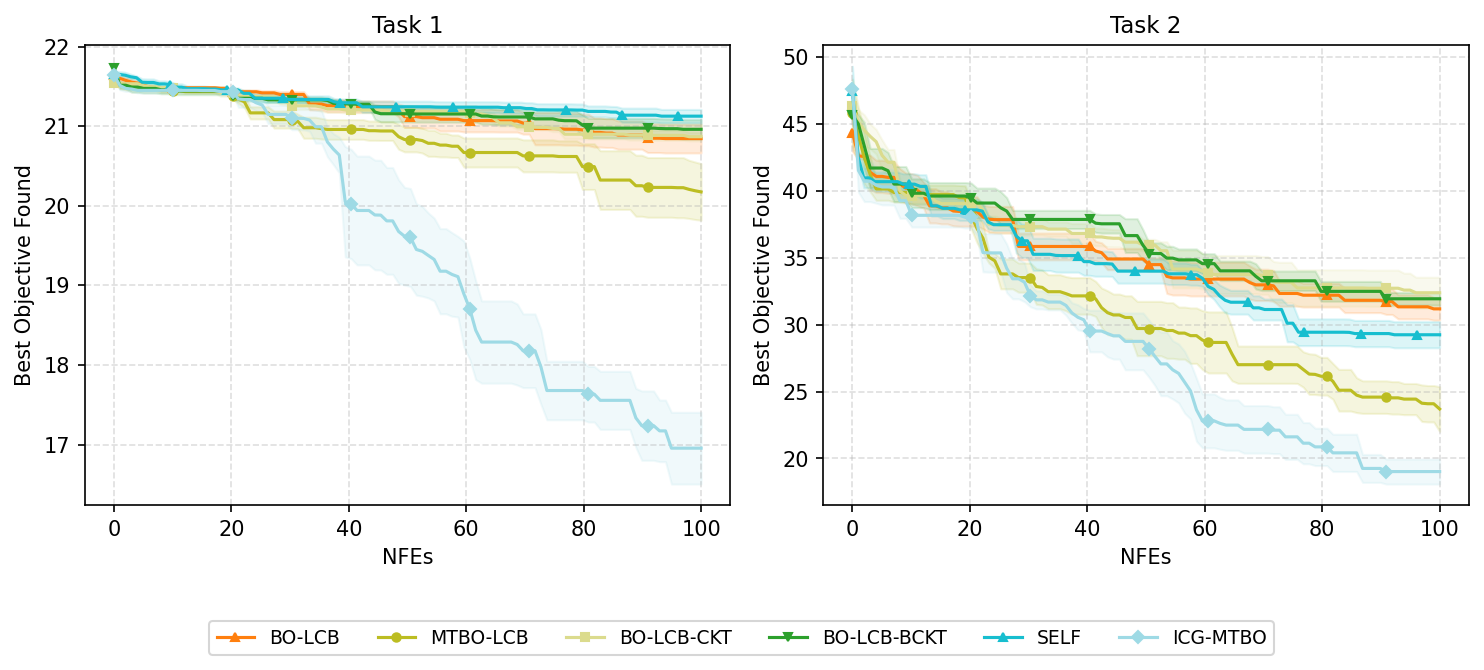}
        \label{fig:ch3-convergence-50d-p2}
    }

    \vspace{0.5mm}

    % Row 2: P7 and P9
    \subfloat[P7 (NIHS)]{
        \includegraphics[width=0.48\textwidth]{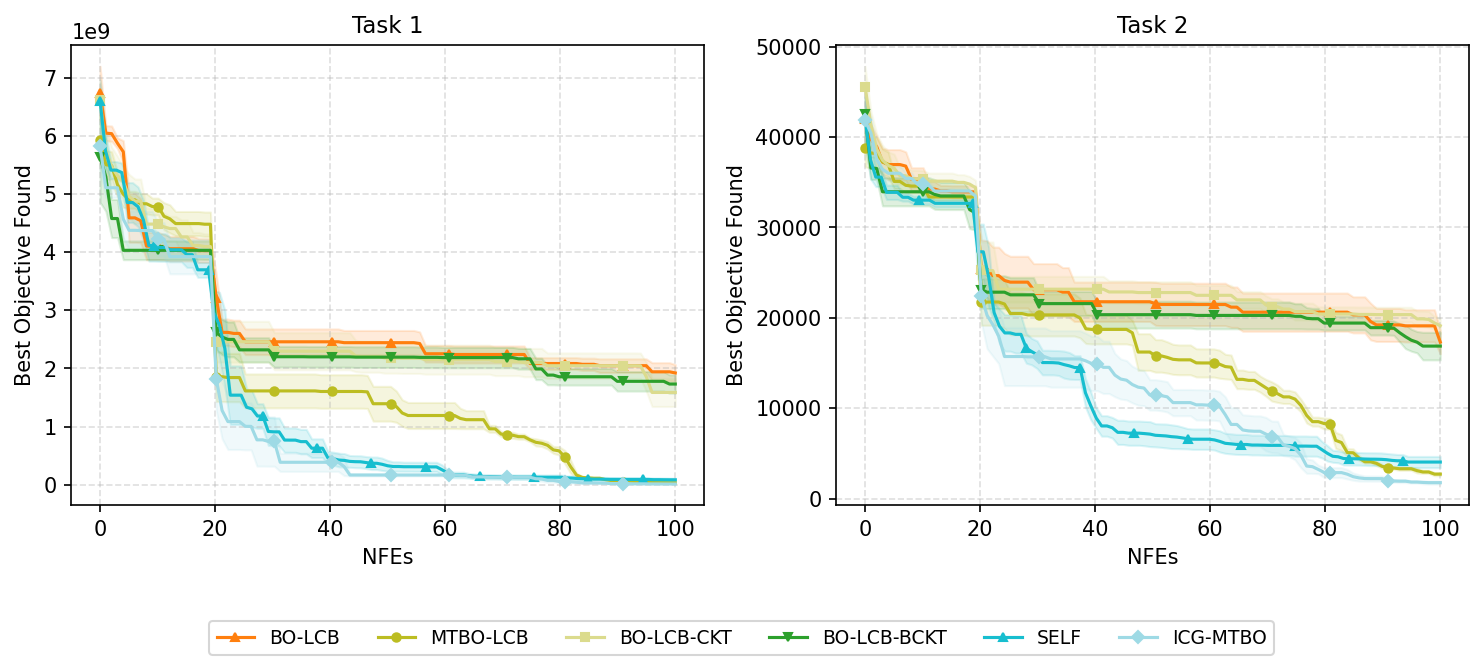}
        \label{fig:ch3-convergence-50d-p3}
    }
    \hfill
    \subfloat[P9 (NILS)]{
        \includegraphics[width=0.48\textwidth]{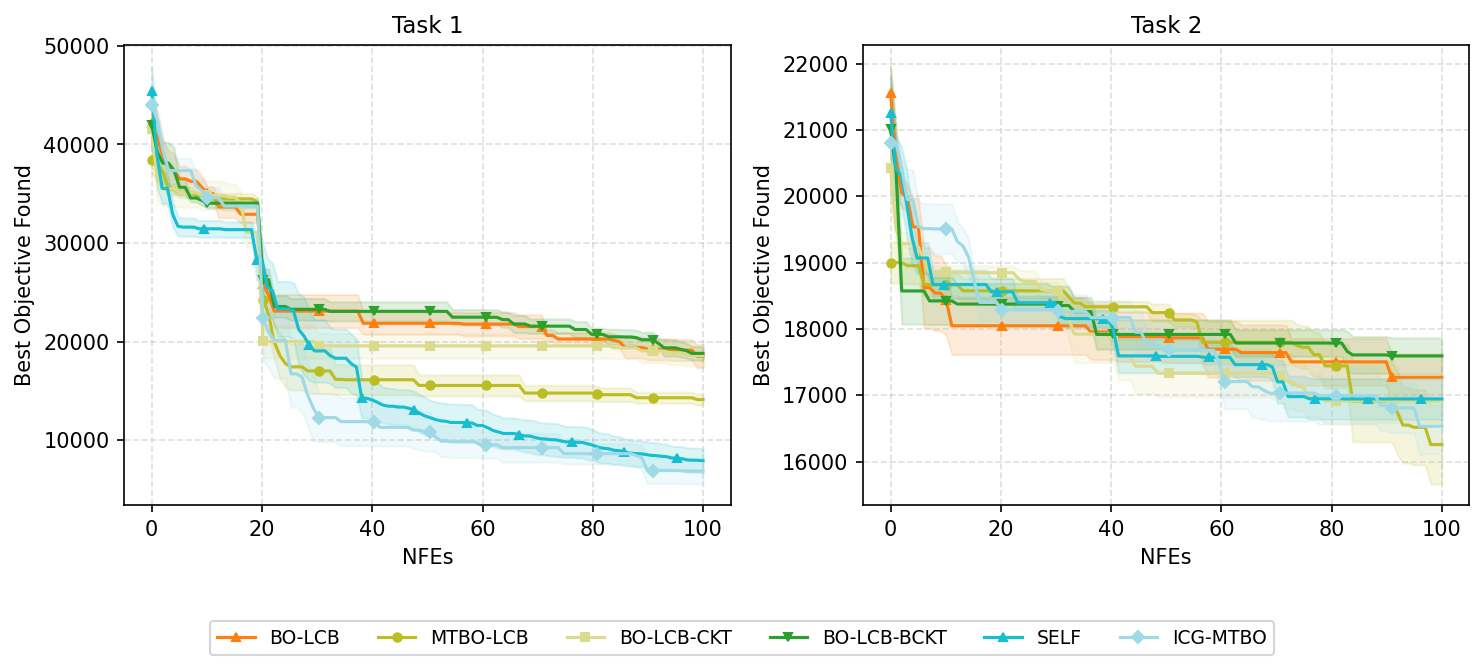}
        \label{fig:ch3-convergence-50d-p4}
    }

    \caption{Convergence trends of BO-LCB, MTBO-LCB, BO-LCB-CKT, BO-LCB-BCKT, SELF, and ICG-MTBO on the 50-dimensional synthetic benchmark problems.
    (a) P1 (CIHS).
    (b) P6 (PILS).
    (c) P7 (NIHS).
    (d) P9 (NILS).}
    \label{fig:ch3-convergence-50d}
\end{figure*}

\begin{table*}[ht]
\centering
\caption{Comparative results of MFEA-II and ICG-MFEA-II with different elite ratios on the synthetic benchmark problems. The results are reported as the mean and standard deviation over 20 independent trials. The symbols $\mathbf{+}$, $\mathbf{-}$, and $\mathbf{\approx}$ indicate that the corresponding method performs significantly better than, significantly worse than, or statistically similarly to ICG-MFEA-II with an elite ratio of $0.1$, respectively, according to the Wilcoxon test at a significance level of $0.05$. The best mean value for each task is highlighted in bold with a shaded background.}
\label{tab:ch3-icg-mfea-elite-ratio}

\resizebox{0.75\textwidth}{!}{%
\begin{tabular}{c|c|c|c|c|c}
\hline
Problem
& Task
& MFEA-II
& \shortstack{ICG-MFEA-II-$(0.5)$}
& \shortstack{ICG-MFEA-II-$(0.3)$}
& \shortstack{ICG-MFEA-II-$(0.1)$}
\\
\hline\hline

% ==================== P1 ====================
\multirow{2}{*}{P1}
& 1
& 1.773e+00 (2.2e-01)~$\mathbf{-}$
& 1.395e+00 (5.0e-02)~$\mathbf{\approx}$
& 1.493e+00 (3.1e-01)~$\mathbf{\approx}$
& \best{1.332e+00 (9.0e-02)}
\\
\cline{2-6}

& 2
& 1.168e+03 (3.1e+02)~$\mathbf{-}$
& 8.741e+02 (5.4e+01)~$\mathbf{-}$
& 8.925e+02 (2.9e+02)~$\mathbf{\approx}$
& \best{7.595e+02 (6.7e+01)}
\\
\hline\hline

% ==================== P2 ====================
\multirow{2}{*}{P2}
& 1
& 1.358e+01 (1.4e+00)~$\mathbf{-}$
& \best{9.976e+00 (9.2e-01)}~$\mathbf{\approx}$
& 1.109e+01 (9.2e-01)~$\mathbf{\approx}$
& 1.115e+01 (6.9e-01)
\\
\cline{2-6}

& 2
& 1.223e+03 (2.4e+02)~$\mathbf{-}$
& \best{8.047e+02 (7.7e+01)}~$\mathbf{\approx}$
& 9.163e+02 (1.6e+02)~$\mathbf{\approx}$
& 9.464e+02 (1.4e+02)
\\
\hline\hline

% ==================== P3 ====================
\multirow{2}{*}{P3}
& 1
& 2.132e+01 (5.2e-02)~$\mathbf{\approx}$
& \best{2.131e+01 (4.3e-02)}~$\mathbf{\approx}$
& 2.135e+01 (4.6e-02)~$\mathbf{\approx}$
& 2.135e+01 (3.4e-02)
\\
\cline{2-6}

& 2
& \best{3.968e+03 (3.2e+02)}~$\mathbf{+}$
& 4.274e+03 (4.1e+02)~$\mathbf{\approx}$
& 4.495e+03 (4.4e+02)~$\mathbf{\approx}$
& 4.482e+03 (1.6e+02)
\\
\hline\hline

% ==================== P4 ====================
\multirow{2}{*}{P4}
& 1
& 1.245e+03 (2.3e+02)~$\mathbf{\approx}$
& 9.289e+02 (2.1e+02)~$\mathbf{\approx}$
& 1.016e+03 (1.5e+02)~$\mathbf{\approx}$
& \best{9.243e+02 (7.2e+01)}
\\
\cline{2-6}

& 2
& 2.809e+03 (7.5e+02)~$\mathbf{\approx}$
& 2.018e+03 (7.9e+02)~$\mathbf{\approx}$
& 2.493e+03 (1.2e+03)~$\mathbf{\approx}$
& \best{1.979e+03 (6.6e+02)}
\\
\hline\hline

% ==================== P5 ====================
\multirow{2}{*}{P5}
& 1
& 1.328e+01 (1.1e+00)~$\mathbf{-}$
& 1.184e+01 (8.4e-01)~$\mathbf{\approx}$
& 1.249e+01 (7.7e-01)~$\mathbf{-}$
& \best{1.141e+01 (6.7e-01)}
\\
\cline{2-6}

& 2
& 2.138e+07 (1.7e+07)~$\mathbf{-}$
& \best{3.039e+06 (1.8e+06)}~$\mathbf{\approx}$
& 5.019e+06 (1.6e+06)~$\mathbf{\approx}$
& 5.355e+06 (3.8e+06)
\\
\hline\hline

% ==================== P6 ====================
\multirow{2}{*}{P6}
& 1
& 1.308e+01 (1.4e+00)~$\mathbf{\approx}$
& 1.298e+01 (1.1e+00)~$\mathbf{\approx}$
& \best{1.267e+01 (7.2e-01)}~$\mathbf{\approx}$
& 1.298e+01 (1.1e+00)
\\
\cline{2-6}

& 2
& 2.479e+01 (2.9e+00)~$\mathbf{\approx}$
& \best{2.452e+01 (1.8e+00)}~$\mathbf{\approx}$
& 2.499e+01 (2.1e+00)~$\mathbf{\approx}$
& \best{2.452e+01 (1.8e+00)}
\\
\hline\hline

% ==================== P7 ====================
\multirow{2}{*}{P7}
& 1
& 1.947e+07 (1.0e+07)~$\mathbf{-}$
& 4.415e+06 (3.1e+06)~$\mathbf{\approx}$
& \best{3.451e+06 (2.5e+06)}~$\mathbf{\approx}$
& 4.830e+06 (3.6e+06)
\\
\cline{2-6}

& 2
& 1.087e+03 (1.7e+02)~$\mathbf{\approx}$
& 1.089e+03 (2.0e+02)~$\mathbf{\approx}$
& \best{1.020e+03 (6.5e+01)}~$\mathbf{\approx}$
& 1.123e+03 (1.5e+02)
\\
\hline\hline

% ==================== P8 ====================
\multirow{2}{*}{P8}
& 1
& 1.751e+00 (1.6e-01)~$\mathbf{\approx}$
& 1.675e+00 (2.2e-01)~$\mathbf{\approx}$
& 1.721e+00 (3.4e-01)~$\mathbf{\approx}$
& \best{1.673e+00 (2.5e-01)}
\\
\cline{2-6}

& 2
& 3.174e+01 (3.1e+00)~$\mathbf{+}$
& \best{3.170e+01 (4.3e+00)}~$\mathbf{\approx}$
& 3.444e+01 (4.8e+00)~$\mathbf{\approx}$
& 3.612e+01 (2.8e+00)
\\
\hline\hline

% ==================== P9 ====================
\multirow{2}{*}{P9}
& 1
& \best{1.063e+03 (7.2e+01)}~$\mathbf{\approx}$
& 1.330e+03 (1.8e+02)~$\mathbf{\approx}$
& 1.268e+03 (2.0e+02)~$\mathbf{\approx}$
& 1.137e+03 (1.9e+02)
\\
\cline{2-6}

& 2
& 4.597e+03 (7.8e+02)~$\mathbf{\approx}$
& 4.263e+03 (4.7e+02)~$\mathbf{\approx}$
& 4.524e+03 (6.1e+02)~$\mathbf{\approx}$
& \best{4.158e+03 (7.0e+02)}
\\
\hline\hline

$+/-/\approx$
& -
& 2/7/9
& 0/1/17
& 0/1/17
& -
\\
\hline

Average rank
& -
& 3.28
& \best{1.83}
& 2.61
& 2.28
\\
\hline\hline

\end{tabular}%
}
\end{table*}

\subsubsection{Results on Real Problems}
As shown in Table~S-IX in supplementary materials, ICG-MTBO achieves the best overall average rank of 1.44, compared with 2.37 for BO-LCB and 2.19 for MTBO-LCB. 
Compared to BO-LCB, ICG-MTBO can achieve significantly better, statistically comparable, and significantly worse results on 8, 18, and 1 out of the 27 tasks, respectively. 
Against the unguided MTBO-LCB, the corresponding results are 7, 19, and 1, indicating that the proposed in-context guidance generally improves or preserves the performance of the underlying multitask optimizer.

The comparison of ICG-MTBO to MTBO-LCB further reveals how the benefit varies with problem dimensionality and inter-task similarity. 
On the 5-dimensional problems from P1 to P3, ICG-MTBO is significantly better on 3 tasks and statistically comparable on the remaining 6. 
On the 10-dimensional problems from P4 to P6, it records 2/6/1 better/comparable/worse results, while on the 15-dimensional problems from P7 to P9, it achieves 2/7/0. 
Thus, the proposed method remains competitive across all three problem scales, although its clearest advantage appears in the lower-dimensional setting. 
More importantly, the improvements mainly revolve around the task group with stronger inter-task relatedness. 
For the HS problems P1, P4, and P7, ICG-MTBO achieves 4/5/0 better/comparable/worse results against MTBO-LCB, while the corresponding result is 3/6/0 for the MS problems P2, P5, and P8. 
In contrast, on the LS problems P3, P6, and P9, the result becomes 0/8/1. 
This trend suggests that the ICG component is most effective when the tasks possess sufficiently overlapping solution structures that can be identified from limited observations. 
When the tasks are weakly related, the guidance provides less additional benefit, but the annealed formulation generally allows ICG-MTBO to remain statistically comparable to standard MTBO-LCB rather than causing persistent negative transfer.

\subsubsection{Generality Study on a Different Acquisition Function}

To examine whether the proposed framework depends on a particular acquisition function, we replace LCB with LogEI while retaining the same surrogate configuration and in-context guidance mechanism. As shown in Tables~S-IV--S-VIII and Tables~III--IV, ICG-MTBO-LogEI achieves the best average rank at all four evaluation budgets, with ranks of 1.17, 1.22, 1.44, and 1.67 after 40, 60, 80, and 100 evaluations, respectively. 
Compared with its unguided counterpart, MTBO-LogEI, the proposed method performs significantly better, statistically similarly, and significantly worse in 37, 26, and 9 out of the 72 comparisons, respectively. At the earliest budget of 40 evaluations, ICG-MTBO-LogEI is better in 10 cases, comparable in 7, and worse in only 1, again demonstrating the value of in-context guidance when the task-coupling estimator is supported by only limited observations.

The advantage gradually decreases as the optimization proceeds: the better/comparable/worse counts change from 10/7/1 at 40 evaluations to 7/8/3 at 100 evaluations.
This trend is consistent with the annealed MAP formulation, where the influence of the in-context guidance is strongest during the early few-shot stage and progressively diminishes as the accumulated observations allow the conventional MTBO estimator to become more reliable. 
Despite this gradual fallback toward standard MTBO-LogEI, the proposed method remains the best-ranked approach throughout the optimization process. 
These results indicate that the benefit of ICG-MTBO is not tied to LCB.
Instead, the proposed framework improves the estimation of inter-task coupling independently of the subsequent acquisition strategy and can therefore be integrated with different Bayesian optimization criteria~\cite{knowledge_gradient, expected_improvement, Human-Loop, aq-total} without modifying their original candidate selection mechanisms.

\subsubsection{Generality Study on an Evolutionary Multitask Optimizer}
\label{sec: ch3_mfeaii}
To demonstrate that the proposed in-context guidance multitask optimization is not tied to the specific multitask Bayesian optimization, we further instantiate the framework on the well-known evolutionary multitask optimizer MFEA-II~\cite{mfea2}. 
In this instantiation, the inter-task relationship takes the form $\boldsymbol{\vartheta}\equiv\mathbf{R}$, where
$\mathbf{R}=[r_{ij}]$ is the random mating probability~(RMP) matrix that governs the intensity of inter-task genetic transfer. 
Conventional MFEA-II estimates $\mathbf{R}$ online per generation from the current parent subpopulations $\mathcal{P}_1,\ldots,\mathcal{P}_M$ by maximizing the likelihood of a mixture model constructed in the unified decision space~\cite{mfea2}:
\begin{equation}
    \widehat{\mathbf{R}}_{\mathrm{MLE}}
    =
    \arg\max_{\mathbf{R}}\;
    \mathcal{L}_{\mathrm{MFEA-II}}
    \left(\mathbf{R};
    \mathcal{P}_1,\ldots,\mathcal{P}_K\right).
\end{equation}
Although this data-driven estimator can enable an adaptive knowledge transfer behavior, it relies on the current populations and can therefore be unreliable in the early generations, when the subpopulations are small and have not yet concentrated to promising regions. 
This corresponds to the same few-shot limitation that motivates the Bayesian instantiation in \ref{sec: ch3-bayesian}.

Following the generic ICG-MTO formulation, an algorithm-specific in-context query can be constructed from the evaluated solutions.
Since the RMP matrix determines whether genetic transfer between the solutions of two tasks is likely to produce valuable offspring, the relevant signal can be modeled as the overlap between their promising regions in the decision space. 
At generation $t$, the elite subset of task $\mathcal{T}_k$ is
defined as:
\begin{equation}
    \mathcal{E}_k^{(t)}
    =
    \operatorname{Elite}_{\alpha}
    \left(\mathcal{P}_k^{(t)}\right),
\end{equation}
where $\gamma$ denotes the elite ratio and the top-$\gamma$ fraction of solutions with the best objective values is retained. 
For each task pair $(\mathcal{T}_i,\mathcal{T}_j)$, the two elite sets are set to the same size and pooled to form a task-membership classification query. 
Each decision vector is labeled according to its original task, while the objective values are used only for elite selection and are not included as input features or predictive labels.

The frozen TabPFN performs in-context classification on a held-out split of this query, thereby serving as a classifier to distinguish samples from the two elite distributions. 
Let $\mathrm{CE}_{ij}^{(t)}$ denote the resulting mean predictive cross-entropy. 
The in-context guidance signal for the task pair is defined as
\begin{equation}
    \rho_{ij}^{(t)}
    =
    \min\left(
    \frac{\mathrm{CE}_{ij}^{(t)}}{\log 2},
    1
    \right)
    \in[0,1].
\end{equation}
A cross-entropy close to $\log 2$ indicates that the two elite sets cannot be reliably distinguished and therefore exhibit substantial decision-space overlap, yielding
$\rho_{ij}^{(t)}\approx1$. 
Conversely, a cross-entropy close to zero indicates well-separated promising regions and yields $\rho_{ij}^{(t)}\approx0$. 
Collecting all pairwise guidance values gives the in-context guidance matrix:
\begin{equation}
    \mathbf{R}_{\mathrm{FM}}^{(t)}
    =
    \left[\rho_{ij}^{(t)}\right]_{i,j=1}^{K}.
\end{equation}

The guided estimator follows the same MAP structure as the generic formulation in Equation~\eqref{eq:ch3-map-general}:
\begin{equation}
    \widehat{\mathbf{R}}_{\mathrm{MAP}}
    =
    \arg\max_{\mathbf{R}}
    \left[
    \mathcal{L}
    \left(
    \mathbf{R};
    \mathcal{P}_1,\ldots,\mathcal{P}_K
    \right)
    -
    \lambda(t)
    \left\|
    \mathbf{R}
    -
    \mathbf{R}_{\mathrm{FM}}^{(t)}
    \right\|^2
    \right].
\end{equation}
Since MFEA-II estimates the RMP independently for each task pair, the implementation solves the scalar problem
\begin{equation}
    \widehat{r}_{ij}^{(t)}
    =
    \arg\max_{r_{ij}\in[0,1]}
    \left[
    \mathcal{L}_{ij}
    \left(
    r_{ij};
    \mathcal{P}_i^{(t)},
    \mathcal{P}_j^{(t)}
    \right)
    -
    \lambda(t)
    \left(
    r_{ij}
    -
    \rho_{ij}^{(t)}
    \right)^2
    \right].
\end{equation}
The pairwise likelihood is normalized by the number of parents so that the guidance weight remains on a scale comparable to that used in the Bayesian instantiation.

The guidance strength follows the same annealing schedule:
\begin{equation}
    \lambda(t)
    =
    \lambda_0\exp(-\delta t).
\end{equation}
The in-context guidance therefore has its greatest influence during the
data-scarce early generations and gradually yields to the
population-based likelihood as the subpopulations mature. As
$\lambda(t)\rightarrow0$, the estimator recovers conventional MFEA-II
exactly. The resulting algorithm is denoted ICG-MFEA-II.

The two variants are compared on the $30$-dimensional synthetic
benchmark suite with a population size of $20$ per task and a budget of $1200$ evaluations per task. To ensure a fair comparison, both methods follow the same initialization protocol. A shared space-filling LHS~\cite{lhs} design of $200$ points per task is evaluated once, the best $20$ solutions form the initial population, and the same evaluated design provides the initial in-context queries for ICG-MFEA-II. Since both methods incur
the same initialization cost, they execute the same number of
generations, namely $50$, and differ only in the estimator of the RMP matrix.

\begin{figure}[h]
\centering
\includegraphics[width=0.35\textwidth]{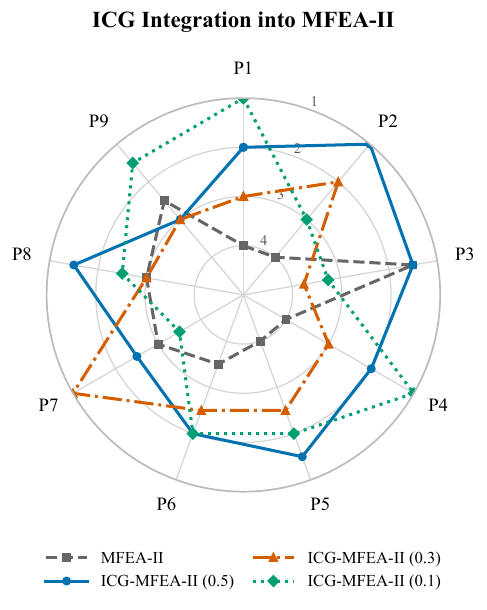}
\caption{Rank-based comparison of MFEA-II and ICG-MFEA-II
with elite ratios of $0.5$, $0.3$, and $0.1$ on the synthetic
benchmark problems P1-P9. For each task, the four methods
are ranked by their mean objective values over 20 independent
trials, with lower values receiving better ranks and ties
receiving average ranks. The radial axis is reversed so rank $1$ is outermost; points farther from the center therefore indicate better relative performance. }
\label{fig:ch3-icg-mfea-elite-ratio-radar}
\end{figure}

The MAP configuration $(\lambda_0=1,\delta=0.05)$ is inherited directly from ICG-MTBO without retuning. The elite ratio is examined at $\gamma\in\{0.5,0.3,0.1\}$, with the corresponding variants denoted ICG-MFEA-II-50, ICG-MFEA-II-30, and ICG-MFEA-II-10, respectively. As shown in Figure~\ref{fig:ch3-icg-mfea-elite-ratio-radar} and Table~\ref{tab:ch3-icg-mfea-elite-ratio}, the guided variants generally improve the convergence of MFEA-II, with the most evident gains appearing in the early generations, when the population-based mixture-model estimator is least reliable.

\subsubsection{Sensitivity Analysis}

The sensitivity of ICG-MTBO to the MAP schedule
$\lambda(t)=\lambda_{0}e^{-\delta t}$, shown in \eqref{eq:ch3-pseudo-observation-decay}, is examined over six configurations as shown in Table S-XI, in the supplementary materials.
In Table~S-XI, the setting $(\lambda_{0}{=}1, \delta{=}0.05)$ attains the best average rank at iteration $60$. 
The detailed analysis and discussion can be found in Section S-II in supplementary materials.

\section{Conclusion}
This paper introduced In-Context Guidance Multitask Optimization (ICG-MTO) for improving inter-task coupling estimation under few-shot evaluation budgets. 
Through the ICG-MTBO instantiation, evaluated solutions were reorganized into directional in-context queries, a frozen numerical foundational model was used to infer predictive inter-task guidance, and the resulting guidance was translated into target-specific reference coupling structures for annealed MAP estimation of the MTGP inter-task coupling matrices. Experimental results on synthetic benchmarks and a real-world problem demonstrated that ICG-MTBO generally achieved better optimization performance than single-task and conventional multitask baselines. 
Its effectiveness across LCB and LogEI, together with the MFEA-II instantiation and sensitivity analysis, further demonstrated the generality and stability of the proposed in-context guidance framework.

Along this line of inquiry, extending in-context guidance to more complex multitask settings, including multiobjective multitask optimization~\cite{moo-finv, wei26} and parametric multitask optimization~\cite{PMTO, wei26}, represents a promising direction for investigating how numerical foundational models can guide inter-task coupling estimation beyond finite single-objective task collections.
% This paper introduced In-Context Guidance Multitask Optimization (ICG-MTO) for improving inter-task relationship estimation under few-shot evaluation budgets. Through the ICG-MTBO instantiation, an in-context query was constructed from the evaluated solutions, a frozen numerical foundational model was used to infer in-context guidance, and the resultant guidance signal was incorporated into the MTGP task kernel through an annealed MAP formulation. 
% The experimental results on synthetic benchmarks and a real-world problem demonstrated that ICG-MTBO generally achieved better optimization performance than single-task, conventional multitask baselines. 
% Its effectiveness across LCB and LogEI, together with the MFEA-II based instantiation and sensitivity analysis, further confirmed the generality and stability of the proposed framework.

% Along the current line of inquiry, more complicated settings including multiobjective multitask optimization~\cite{moo-finv, wei2025amortized} and parametric multitask optimization~\cite{PMTO, wei2025amortized} are worth further investigation with foundational models involved.

\bibliographystyle{IEEEtran}
\bibliography{bibfiles_async}

\end{document}